\documentclass{article} % For LaTeX2e
\usepackage{iclr2025_conference,times}

\usepackage{hyperref}
\usepackage{url}
\usepackage{graphicx}
\usepackage{caption}
\usepackage{wrapfig}
\usepackage{subcaption}
\usepackage{enumitem}
\usepackage{comment}
\usepackage{amssymb}
\usepackage{booktabs}
\usepackage{algorithm} 
\usepackage{algpseudocode}  
\usepackage{amsmath,amsfonts,bm}
\usepackage{amsthm}
\usepackage{mathtools}
\usepackage{xspace}
\usepackage{cleveref}
\usepackage{bbold}
\usepackage{url}

\title{Grokking at the Edge of Numerical Stability}

\author{\textbf{Lucas Prieto},\, \textbf{Melih Barsbey},\, \textbf{Pedro A.M. Mediano\thanks{Joint senior authors, equal contribution}} ,\, \textbf{Tolga Birdal}\footnotemark[1] \\
Department of Computing\\
Imperial College London\\
}

\def\eqref#1{equation~\ref{#1}}
\def\1{\bm{1}}

\DeclareMathAlphabet{\mathsfit}{\encodingdefault}{\sfdefault}{m}{sl}
\SetMathAlphabet{\mathsfit}{bold}{\encodingdefault}{\sfdefault}{bx}{n}

\newcommand{\tolga}[1]{\textcolor{blue}{T\string: #1}}

\newcommand{\nlm}{NLM\xspace}

\newcommand{\loss}{\mathcal{L}}

\newcommand{\ograd}{\ensuremath{\perp\!\mathrm{\!Grad}}\xspace}

\newcommand{\softmax}{\ensuremath{\mathrm{Softmax}}\xspace}
\newcommand{\tsoftmax}{\ensuremath{\mathrm{Taylor-Softmax}}\xspace}
\newcommand{\bfsoftmax}{\mathbf{Softmax}}
\newcommand{\stablemax}{\mathrm{StableMax}}

\newcommand{\params}{\boldsymbol{\theta}}

\crefname{equation}{eq.}{eq.}
\Crefname{equation}{Eq.}{Eq.}
\crefname{theorem}{thm.}{thms.}
\Crefname{Theorem}{Thm.}{Thms.}
\crefname{proposition}{prop.}{props.}
\Crefname{proposition}{Prop.}{Props.}
\crefname{definition}{dfn.}{dfn.}
\Crefname{definition}{Dfn.}{Dfn.}
\crefname{remark}{remark}{remark}
\Crefname{Remark}{Remark}{Remark}
\crefname{algorithm}{Alg.}{Alg.}
\Crefname{Algorithm}{Alg.}{Alg.}

\newtheorem{prop}{Proposition}
\newtheorem{dfn}{Definition}

\newcommand{\ie}{\textit{i}.\textit{e}. }

\crefname{section}{Sec.}{Secs.}
\Crefname{section}{Sec.}{Secs.}
\crefname{equation}{Eq.}{Eqs.}
\Crefname{equation}{Eq.}{Eqs.}
\crefname{figure}{Fig.}{Figs.}
\Crefname{figure}{Fig.}{Figs.}
\crefname{table}{Tab.}{Tabs.}
\Crefname{table}{Tab.}{Tabs.}
\crefname{thm}{Thm.}{Thms.}
\Crefname{thm}{Thm.}{Thms.}
\crefname{dfn}{Dfn.}{Dfns.}
\crefname{dfn}{Dfn.}{Dfns.}
\crefname{remark}{remark}{remarks}
\Crefname{Remark}{Remark}{Remarks}
\crefname{prop}{Prop.}{Prop.}
\Crefname{prop}{Prop.}{Prop.}
\Crefname{algorithm}{Alg.}{Alg.}
\crefname{appendix}{App.}{apps.}
\Crefname{appendix}{App.}{Apps.}
\crefname{appsec}{appendix}{appendices}
\Crefname{appsec}{Appendix}{Appendices}

\renewcommand{\paragraph}[1]{{\vspace{0.5mm}\noindent \bf #1}.}

\newcommand\figsubref[2]{\hyperref[#1]{\ref*{#1}#2}}

\iclrfinalcopy
\begin{document}

\maketitle
 
\begin{abstract}
Grokking, or sudden generalization that occurs after prolonged overfitting, is a surprising phenomenon that has challenged our understanding of deep learning. While a lot of progress has been made in understanding grokking, it is still not clear why generalization is delayed and why grokking often does not happen without regularization. In this work we argue that without regularization, grokking tasks push models to the edge of numerical stability, introducing floating point errors in the Softmax that we refer to as \emph{Softmax Collapse} (SC). We show that SC prevents grokking and that mitigating SC leads to grokking \textit{without} regularization. Investigating the root cause of SC, we find that beyond the point of overfitting, the gradients strongly align with what we call the \emph{naïve loss minimization} (NLM) direction. This component of the gradient does not change the predictions of the model but decreases the loss by scaling the logits, usually through the scaling of the weights along their current direction. We show that this scaling of the logits explains the delay in generalization characteristic of grokking, and eventually leads to SC, stopping learning altogether. To validate these hypotheses, we introduce two key contributions that mitigate the issues faced in grokking tasks: (i) $\stablemax$, a new activation function that prevents SC and enables grokking without regularization, and (ii) $\ograd$, a training algorithm that leads to quick generalization in grokking tasks by preventing NLM altogether. These contributions provide new insights into grokking, shedding light on its delayed generalization, reliance on regularization, and the effectiveness of known grokking-inducing methods. Code for this paper can be found at: \url{https://github.com/LucasPrietoAl/grokking-at-the-edge-of-numerical-stability}.

\end{abstract}

\section{Introduction}

Deep learning has been transformative for a variety of fields such as natural language processing~\citep{devlin2018bert}, computer vision~\citep{krizhevsky2012imagenet}, geometry processing~\citep{qi2017pointnet}, and 3D vision~\citep{deng2018ppfnet}. This rapid proliferation has brought with it surprising phenomena that defy the predictions of classical statistical learning theory.

In this paper we explore one such recently observed phenomenon known as \emph{grokking}, first described by \citet{power2022grokking} as a sudden and unexpected generalization occurring after prolonged overfitting. Although predominantly studied in algorithmic tasks like modular addition or multiplication, recent findings suggest that grokking may be a more pervasive phenomenon, also manifesting in more complex tasks involving vision and language~\citep{lv2024language,humayun2024deep}.

Prior research has consistently observed grokking in settings that involve some form of regularization, such as weight decay~\citep{Barak2022-el, power2022grokking, Nanda2023-hf}. This pattern has motivated investigations into the implicit biases introduced by weight decay, suggesting it may be critical to triggering delayed generalization. For instance, \citet{liu2023omnigrok} argued that weight norms need to be in a narrow range or ``Goldilocks Zone'' for generalization. Similarly, \citet{Varma2023} highlighted weight efficiency of generalizing solutions, and \citet{Nanda2023-hf} argued that weight decay favors simpler, more generalizable solutions. However, recent works have argued that regularization may not be necessary for grokking, at least on shallow networks with Mean Squared Error (MSE) loss \citep{Kumar2023-hz, Lyu2023-ga, Gromov2023-nh}. These works tie grokking to a transition from lazy training \citep{Chizat_Oyallon_Bach_2018} to feature learning. Despite this ongoing work, several aspects in this framing of grokking remain unclear. These include why grokking tasks induce lazy training and why weight decay is often needed to enter the feature learning regime when using deeper models or cross-entropy (CE) loss.

\begin{figure}[t]
\begin{centering}
    \includegraphics[width=\linewidth]{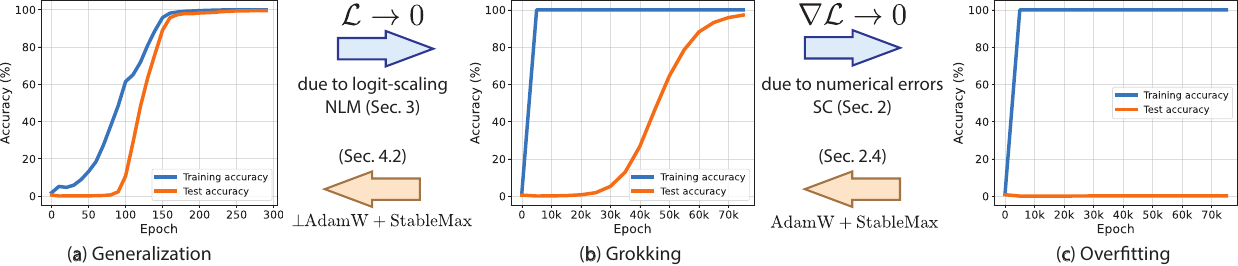}\vspace{-6mm}
\end{centering}
\caption{Our contributions demonstrated through results obtained in addition modulo 113 task. We show that the delay in generalization induced by NLM can be reversed using the proposed $\perp$\!AdamW ((\textbf{a}) and (\textbf{b})) and that the numerical errors that lead to overfitting instead of grokking can be avoided by using the proposed $\stablemax$ ((\textbf{b}) and (\textbf{c})). \vspace{-5mm}}
\label{fig:teaser}
%\vspace{-3mm}
\end{figure}

Here we propose a novel account of grokking, outlined in \cref{fig:teaser}, that explains several of the main unanswered questions in the grokking literature. We start by showing that without regularization, grokking is prevented by absorption errors in the \softmax, which we call \emph{Softmax Collapse} (SC). These errors result in zero terms in the gradient and put an end to learning, sometimes before any progress is made in the test performance, resulting in complete overfitting (\cref{fig:teaser}, \textbf{c}). We then argue that SC is caused by what we call \emph{Naïve Loss Minimization} (NLM), as the gradient becomes aligned with a direction that corresponds to scaling up the logits by a constant. While scaling up all the logits does not change the model predictions, it does reduce the CE loss for a network that has reached 100\% training accuracy, with the downside that this eventually leads to numerical errors in \softmax. 
Our findings provide explanations for several key aspects of grokking, including (i) the delayed onset of generalization, (ii) why grokking is often absent without regularization, and (iii) why existing methods designed to induce grokking are effective.

To validate our hypothesis that SC is responsible for the absence of grokking without regularization, we introduce $\bm{\stablemax}$ as a more numerically stable replacement to $\softmax$ in CE loss. This simple change takes models from complete overfitting to grokking (\cref{fig:teaser}, \textbf{c} to \textbf{b}) \textit{without} regularization, in settings where it is normally not observed without it. Similarly, we validate that NLM is responsible for delaying generalization (\cref{fig:teaser}, \textbf{a} to \textbf{b}) and leading to SC by introducing a new optimizer $\ograd$, which only preserves the part of the gradient that is orthogonal to the NLM direction. By doing this, $\ograd$ quickly leads to generalization without the initial overfitting phase that defines grokking (\cref{fig:teaser}, \textbf{b} to \textbf{a}).

Our primary contributions are as follows:
\begin{itemize}[leftmargin=*,topsep=0em,noitemsep]
    \item We observe that cases of overfitting without grokking are due to floating point errors caused by extreme values in the $\softmax$~function, which we term Softmax Collapse (SC;~\cref{sec:floating_points}).
    \item We show that interventions to avoid SC, like greater floating point precision or a new, numerically stable version of Softmax ($\stablemax$), cause grokking in settings where it was previously absent without regularization (\cref{sec:preventing_sc}).
    \item We observe that models move towards SC because overfitting and cross-entropy loss push the model in a direction of uncontrolled logit growth, which we refer to as Naïve Loss Minimization (NLM;~\cref{sec:nlm}).
    \item We demonstrate that NLM can be avoided through a novel optimizer, \ograd, which removes the delay in generalization (\cref{sec:avoiding_nmm}).
\end{itemize}

\begin{comment}
\begin{figure}[t]
\begin{subfigure}[t]{.32\textwidth}
    \includegraphics[width=\linewidth]{grokking_iclr/figures/teaser_baseline.pdf}
    \caption{AdamW}
\end{subfigure}
\hfill
\begin{subfigure}[t]{.32\textwidth}
    \includegraphics[width=\linewidth]{grokking_iclr/figures/teaser_softermax.pdf}
    \label{fig:input_representations}
    \caption{AdamW + stablemax}
\end{subfigure}%
\hfill
\begin{subfigure}[t]{.32\textwidth}
    \includegraphics[width=\linewidth]{grokking_iclr/figures/teaser_nlm.pdf}
    \label{fig:gradient_norms}
    \caption{$\perp$AdamW + stablemax}
\end{subfigure}
\caption{\vspace{-5mm}}
\vspace{-5mm}
\end{figure}
\end{comment}
\section{Setup}
\subsection{Datasets}

We show our findings on the most commonly studied grokking datasets, outlined in this section.

\begin{comment}
    \begin{enumerate}[itemsep=1mm,topsep=0mm,leftmargin=*]
    \item[\textbf{I. Modular Arithmetic}] 
\end{enumerate}
\end{comment}

\paragraph{I. Modular arithmetic}
The main results in this paper are shown on arithmetic modulo 113 \citep{power2022grokking, Nanda2023-hf}. This is a family of supervised learning tasks where two one-hot encoded inputs representing integers $a,b < p$ are used to predict the target $y=a*b \mod p$, where $*$ is some binary operation and $p$ is a prime number. In most of our results, the binary operation is addition, but we show additional results with multiplication and subtraction. 

Modular arithmetic tasks are characterized by a binary operation and a dataset size, with different behaviors being observed for different dataset sizes on the same binary operation. In these settings, we describe the dataset sizes as the percentage of the $113^2$ possible pairs that are used for training, with the rest of the data being used for testing as in \cite{Nanda2023-hf} and \cite{power2022grokking}. Our main results use a 40\%/60\% train/test split but we also include results using 60\%/40\% and 70\%/30\%. The input integers are represented as one-hot vectors. 

\paragraph{II. Sparse parity}
We also validate some of our results on the Sparse Parity task outlined in \cite{Barak2022-el}. This is a supervised learning setting where the target is the parity of $k$ bits out of a binary vector of length $n$, with $k\ll n$. In this work we use 2000 samples, split evenly between train and test data and we describe instances of this task by specifying the values of $n$ and $k$.

\paragraph{III. MNIST}
Finally, we provide some results on a subset the classic image classification dataset MNIST \citep{deng2012mnist}. For our experiments, we use a subset of 200 training samples from the training set as in \cite{liu2023grokking}, with evaluation on the full test set.

\subsection{Models}
 We study the grokking phenomenon on these datasets using a 2-hidden layer multi-layer perceptron (MLP) of width 200 as in \cite{liu2023omnigrok} and a one-layer transformer with 4 attention heads as \cite{Nanda2023-hf} and \cite{power2022grokking}. We train both of these models in a full batch setting, using ReLU activations and cross-entropy loss with AdamW and SGD, as well as our own variants of these optimizers, $\perp$AdamW and $\perp$SGD. Unless specified otherwise we set the weight decay parameter $\lambda=0$. For modular arithmetic datasets, inputs are concatenated as the input of the MLP resulting in a 226 dimensional vector, and treated as separate tokens in the case of the transformer.

\section{Softmax Collapse: Floating Point Errors Prevent Grokking} \label{sec:floating_points}

Given our current understanding of grokking, it is surprising that it happens without regularization for some dataset sizes, but regularization becomes crucial as dataset size decreases \citep{power2022grokking}. In this section we highlight that looking at datasets at the boundary of these two regimes reveals that without weight decay, grokking sometimes starts before abruptly stopping (\cref{fig:grokking_stops}). We show that this is caused by floating point errors in the \softmax that lead the gradients from a large fraction of the samples to become zero. We refer to this phenomenon as Softmax Collapse. 

\subsection{Softmax Collapse}

In modern neural network implementations, Floating Point (FP) arithmetic is ubiquitous for representing and computing parameters, activations, and gradients. While FP numbers enable efficient decimal computations, they introduce numerical inaccuracies. This section focuses on \textit{absorption errors}, as a specific class of FP arithmetic failure. We will use the symbol $\doteq$ to refer to equality under FP arithmetic.

\begin{dfn}[Absorption Errors]
Let $a, b \in \mathbb{R}\setminus\{0\}$ be floating point numbers in a system with base $\beta$ and $p$ significand bits. Denote their exponents by $e_a$ and $e_b$, respectively. An \emph{absorption error} occurs in the computation of $a + b$ (denoted $a + b \doteq a$) if
\[
e_a - e_b \geq p.
\]
In this case, after exponent alignment, the significand of $b$ is shifted right by at least $p$ digits, and $b$ cannot be represented in the available precision, resulting in $a + b \doteq a$.
\end{dfn}

Intuitively, absorption errors can occur during FP addition when operands have significantly different magnitudes. For $float32$ the base $\beta$ is 2 and $p=24$ bits, meaning that adding any number smaller than $2^{-(p-1)}=2^{-23} $ to 1 will leave 1 unchanged. $2^{-23}$ is the machine epsilon for float32.

\paragraph{Absorption errors in the $\bfsoftmax$}
The $\softmax$ function is a fundamental component in numerous deep learning architectures, serving as an activation function or a key element in attention mechanisms. In this case, we focus on its application within the Softmax Cross-Entropy (SCE) loss:

\begin{dfn}[Softmax Cross-Entropy (SCE) loss]
    For a neural network $f$ and a data point $\textbf{x}$ with label $y$, we define $\textbf{z} \coloneq f(\textbf{x})$ and $z_y$ as the logit corresponding to the true class $y$ . We express the SCE loss as well as its equivalent numerically more stable formulation as:
\begin{equation}
    \loss_{\mathrm{SCE}}(f(\textbf{x}), y) = -\log\left(\frac{e^{z_y}}{\sum_{k=1}^n e^{z_k}}\right)= -z_y + \max(\textbf{z}) + \log\left(\sum_{k=1}^n e^{z_k-\max(\textbf{z})}\right)
    \label{eq:log_softmax}
\end{equation}
\end{dfn}

Unfortunately, even the rightmost (comparatively more stable) variant does not address this problem, since the kind of FP errors discussed in this work appear in the sum. While the \softmax function outputs are bounded between 0 and 1, the intermediate calculations involve summing exponentials of both positive and negative logits. These values can span several orders of magnitude, particularly in scenarios with large logits where the loss approaches zero. This wide range of values creates conditions that lead to absorption errors -- leading to the phenomenon we call \textit{Softmax Collapse}.

\begin{dfn}[Softmax Collapse (SC)]
    A specific case of absorption error occurs when, for a given sample $\textbf{x}$, the logit from the correct class $z_y$ is significantly larger than the logits for all other classes. This floating-point absorption of smaller terms, which we call \textbf{Softmax Collapse}, occurs when:
    \begin{equation}
    \sum_{k=1}^n e^{z_k} \doteq e^{z_y} ~ ,
    \label{eq:softmax_collapse}
    \end{equation}
    in which case the SCE loss becomes:
    \begin{equation}
        \loss_{\mathrm{SCE}}(f(\textbf{x}), y) \doteq -\log\left(\frac{e^{z_y}}{e^{z_y}}\right) = 0 ~ .
    \end{equation}
\end{dfn} 

Thus, during SC the loss becomes identical to zero. Furthermore, for the correct class, the gradients become zero as well: 
\begin{equation}
    \frac{\partial{ \loss_{SCE}}}{\partial z_{c}} =  \frac{e^{z_c}}{\sum_{k=1}^n e^{z_k}}-\mathbb{1}_{\{c = y\}} \doteq 1 - \mathbb{1}_{\{c = y\}} ~ .
    \label{eq:derivative}
\end{equation}
While weights that contribute to the wrong classes can still get negative updates, we show that disappearance of the gradients from the correct classes is enough to inhibit grokking (\cref{fig:grokking_stops}). We validate this in \cref{app:sc_intervention} with an explicit intervention, showing that artificially setting the gradients from the correct class to zero stops generalization in a very similar way to what we observe in \cref{fig:grokking_stops}.

\subsection{Evidence of Softmax Collapse in grokking tasks}

\begin{figure}[t]
\centering
\begin{subfigure}{.32\textwidth}
  \centering
  \includegraphics[width=\linewidth]{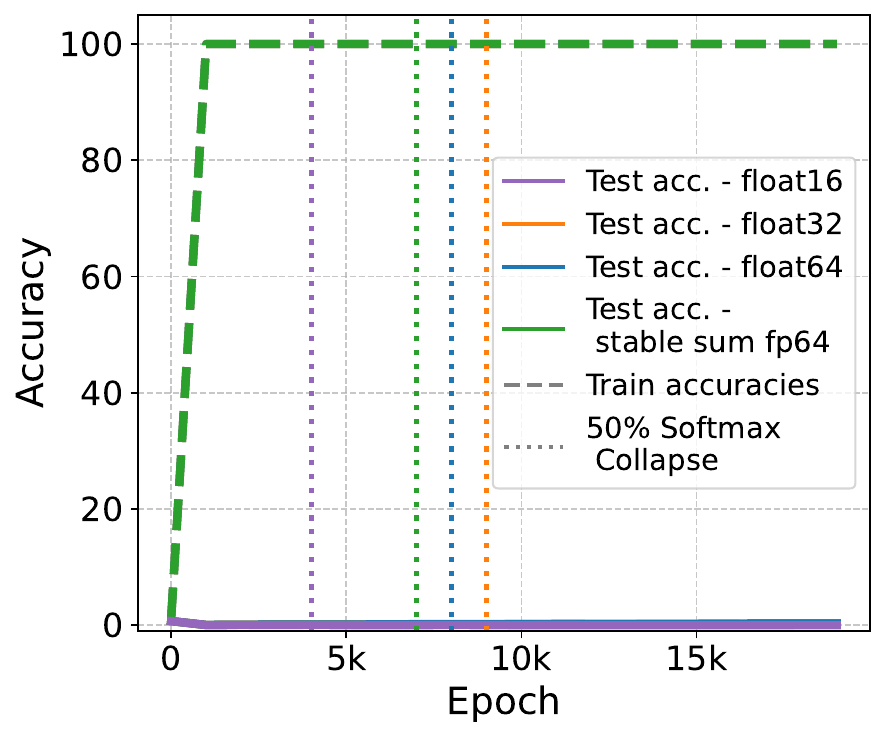}
  \caption{40\% training data}
  \label{fig:grokking_stops_40}
\end{subfigure}
\hfill
\begin{subfigure}{.32\textwidth}
  \centering
  \includegraphics[width=\linewidth]{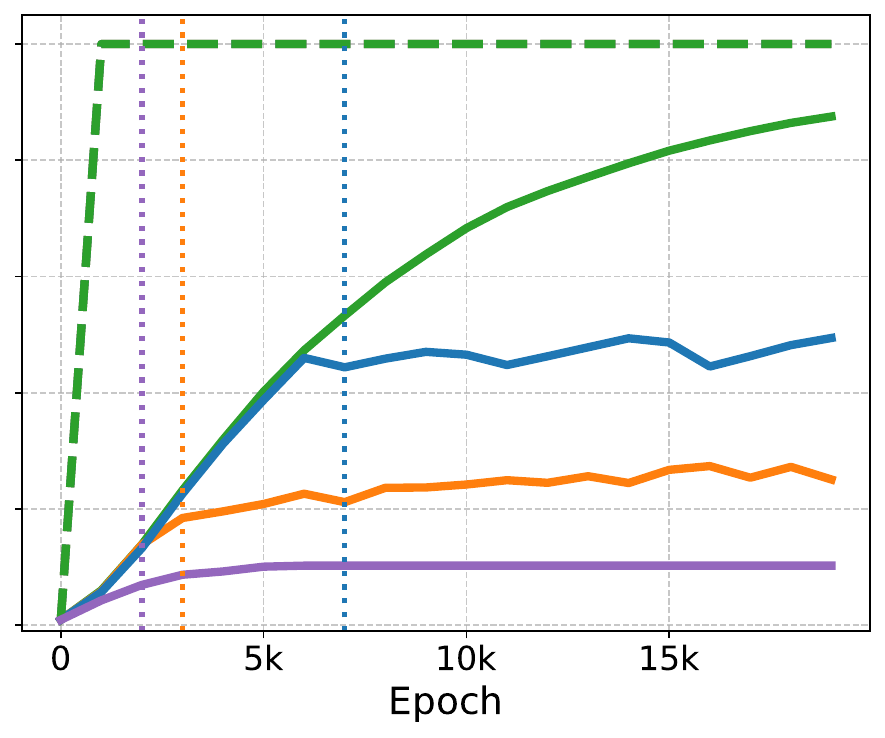}
  \caption{60\% training data}
  \label{fig:grokking_stops_60}
\end{subfigure}
\hfill
\begin{subfigure}{.32\textwidth}
  \centering
  \includegraphics[width=\linewidth]{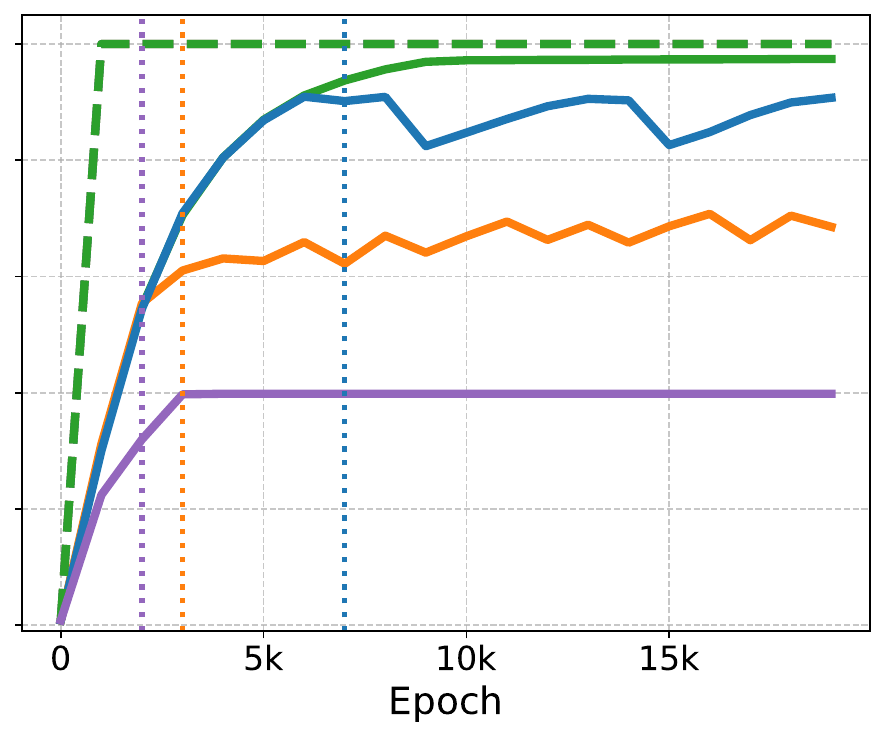}
  \caption{70\% training data}
  \label{fig:grokking_stops_70}
\end{subfigure}
% \vspace{-3mm}
\caption{As dataset size increases (subplots \textbf{a} to \textbf{c}), MLPs trained on modular addition begin to generalize without regularization until this is stopped by SC making the gradient from a large fraction of the samples equal to zero. This stopping point comes earlier for $\mathrm{float32}$ than $\mathrm{float64}$ and with small enough datasets it comes before the model makes any progress on test accuracy.\vspace{-3mm}}
\label{fig:grokking_stops}
\end{figure}
%\tolga{why not have the legend in the first figure? this seems to waste space again.}\lucas{I thought putting the legend in one of the three figures was not commonly done, I can put it in the first one if you think it looks better}}

\label{sec:evidence-of-sc}
Grokking is often studied using dataset sizes for which the delay in generalization is significant, which is usually when the dataset is small but just large enough that generalization is possible. In this regime, regularization seems necessary for grokking and no improvement in test performance is observed without it~\citep{Nanda2023-hf}. However, a fact that has received less attention is that grokking can happen without regularization if the dataset is large enough~\citep{power2022grokking}.

Here we hypothesize that as the size of the dataset decreases, overfitting becomes easier and Softmax Collapse (SC) happens earlier. To quantify this, we train an MLP without regularization on modular addition using different levels of FP precision, and calculate at every training epoch the fraction of samples that result in SC as per \cref{eq:softmax_collapse}. 
The results support our hypothesis that SC is responsible for the model's failure to generalize (\cref{fig:grokking_stops}). Specifically, we see that generalization stops when SC begins -- and that this happens earlier under $\mathrm{float32}$ than under $\mathrm{float64}$ (\cref{fig:grokking_stops_60}). Furthermore, this point is reached earlier as the dataset size decreases until it is reached before making any progress in the test accuracy, resulting in the common picture of no grokking without regularization (\cref{fig:grokking_stops_40}).

\subsection{Preventing Softmax Collapse leads to grokking}
\label{sec:preventing_sc}

To validate the importance of FP errors in stopping grokking, we show that methods to avoid SC lead to generalization on all the common grokking tasks on both MLPs and transformers. We introduce the following methods to postpone the appearance of FP errors.

\paragraph{Increasing floating point precision}
The simplest way to avoid SC is to extend the FP precision from $\mathrm{float32}$ to $\mathrm{float64}$ for the \softmax calculation. We see in~\cref{fig:grokking_stops} that networks trained using $\mathrm{float64}$ in the \softmax face SC later in training which allows for a further increase in test performance. Conversely, using $\mathrm{float16}$ leads to SC earlier in training, leading to lower test performance.
While this approach works as expected, FP precision cannot be extended indefinitely to allow for generalization as seen in the lack of grokking in \cref{fig:grokking_stops_40}.

\paragraph{$\bm{\stablemax}$ Cross Entropy (StCE) Loss} As demonstrated above, SC is caused by adding the exponentials of very large positive and negative logits in the $\softmax$. To avoid these extreme summands, we propose using a softer version of $\softmax$ to transform logits into probabilities before calculating the CE Loss: %Our proposed $\stablemax$ function can be viewed as applying the $g$ function before a regular $\softmax$, or a modified $\softmax$ using the $s$ function instead of an exponential:
% \tolga{Also, what is a $g$-function?}

\begin{wrapfigure}[10]{R}{0.275\textwidth}
            \vspace{-6.mm}
            \begin{center}
    \includegraphics[width=0.275\textwidth]{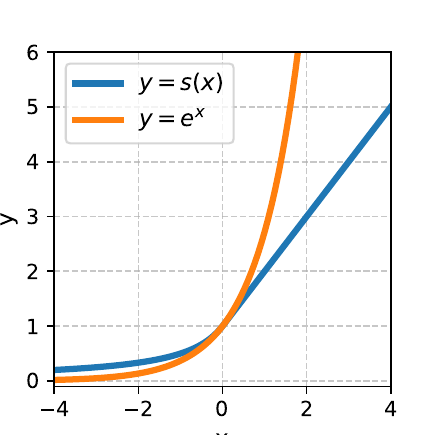}
            \end{center}
            \vspace{-18pt}
            \caption{$s(x)~\mathrm{vs.}~e^x$.}
            \label{fig:stablemax}
        \end{wrapfigure} 
        \noindent\textbf{~~}\vspace{-2.mm}
\begin{dfn}[$\stablemax$]
We introduce a numerically stable version of the \softmax as: 
\begin{equation}
    \stablemax(x_i) \coloneq \frac{s(x_i)}{\sum\limits_j s(x_j)},
\end{equation}
where
\begin{equation}
s(x) \coloneq \begin{cases}
x+1 & \text{if } x \geq 0, \\
\frac{1}{1-x} & \text{if } x < 0
\end{cases}.
\end{equation}
\end{dfn}

As seen in~\cref{fig:stablemax}, $s(\cdot)$ is a simple ramp function that scales linearly instead of exponentially when $x\geq0$ and also approaches 0 more slowly than the exponential function when $x<0$. This is similar to the Softplus function \citep{softplus} but approaches 0 more slowly with negative logits, further reducing the risk of absorption errors.

\begin{prop}\label{prop:stablemax}
$\stablemax$ is a modified $\softmax$, \ie 
$\stablemax\left(x_i\right) = \softmax\left(g\left(x_i\right)\right)$ where
\begin{equation}
         g(x) = \begin{cases}
\log(x+1) & \text{if } x \geq 0, \\
-\log(-x+1) & \text{if } x < 0
\end{cases}.
\end{equation}
\end{prop}

The proof of this Proposition is presented in~\cref{app:proofs}.
We then define the numerically stable analogue of $\loss_{\mathrm{SCE}}$ as $\loss_{\mathrm{StCE}}(f(\textbf{x}), y) = -\log(\stablemax(z_y))$, where $z_y$ again corresponds to the logit of the true class $y$.

\begin{comment}
    \begin{equation}
g(x) = \begin{cases}
\log(x+1) & \text{if } x \geq 0, \\
-\log(-x+1) & \text{if } x < 0
\end{cases}\qquad 
s(x) = \begin{cases}
x+1 & \text{if } x \geq 0, \\
\frac{1}{1-x} & \text{if } x < 0.
\end{cases}
\end{equation}

\begin{equation}
s(x) = \begin{cases}
x+1 & \text{if } x \geq 0, \\
\frac{1}{1-x} & \text{if } x < 0.
\end{cases}
\end{equation}
\end{comment}

\begin{figure}[t]
\begin{subfigure}[t]{.32\textwidth}
    \includegraphics[width=\linewidth]{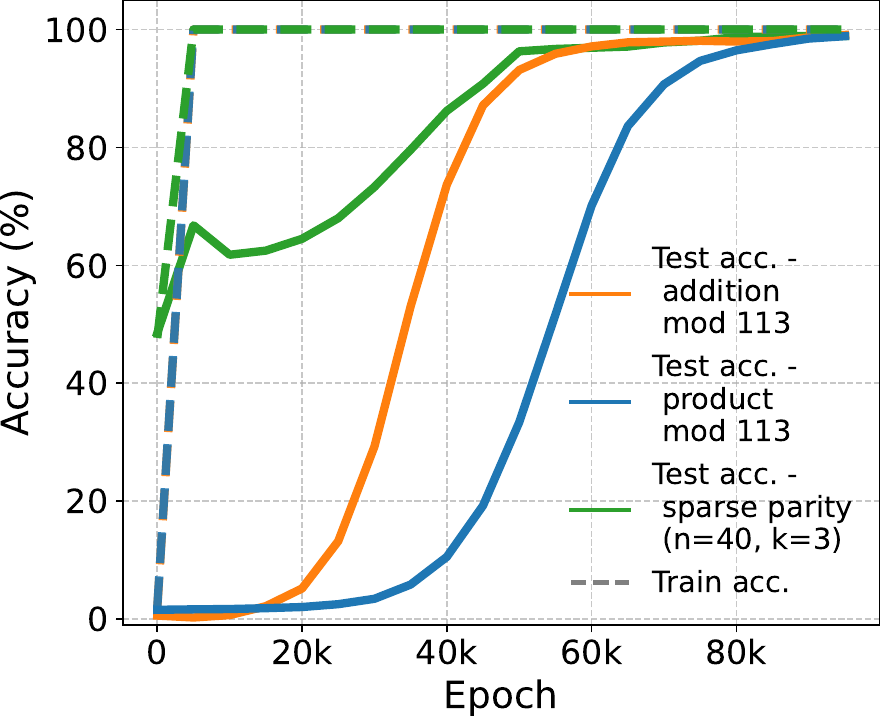}
    %\caption{}
    \label{fig:stablemax_grokking}
\end{subfigure}
\hfill
\begin{subfigure}[t]{.32\textwidth}
    \includegraphics[width=\linewidth]{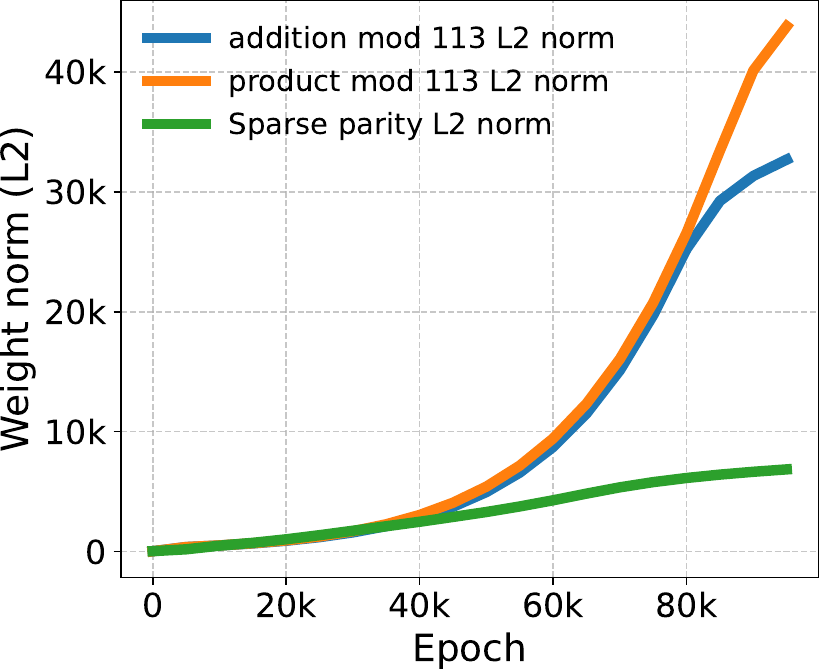}
    %\caption{}
    \label{fig:weight_norms}
\end{subfigure}
\hfill
\begin{subfigure}[t]{.32\textwidth}
    \includegraphics[width=\linewidth]{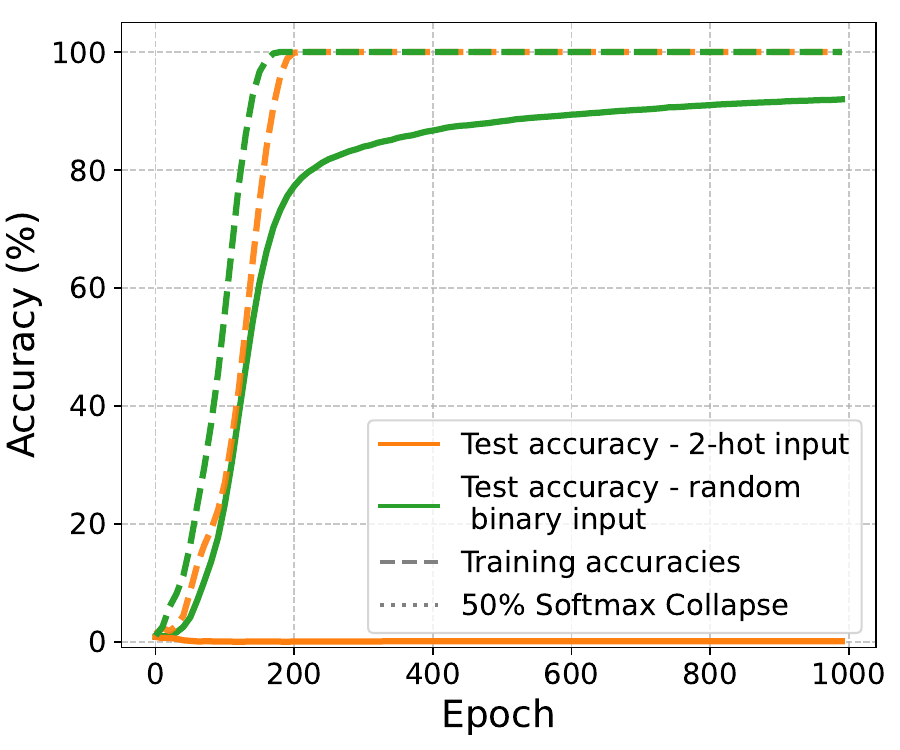}
    %\caption{}
\end{subfigure}%
\vspace{-5mm}
    \caption{(\textbf{left}) Grokking with StCE loss and no regularization on three common grokking datasets using an MLP with 2 hidden layers of width 200. We use  40\% of all pairs modulo 113 which is the same setting as \cref{fig:grokking_stops_40} where regular SCE gets stuck at random level performance (random level is 50\% for sparse parity). (\textbf{middle}) Evolution of model weight norms during training for the same models and tasks. This shows that grokking induced without weight decay does not follow the commonly observed trend of rapidly decreasing weight norm during generalization. (\textbf{right}) Changing input representations turns modular addition into regular machine learning tasks with train and test accuracy increasing in tandem, see \cref{sec:nlm}.\vspace{-5mm}}
    \label{fig:input_representations}
\end{figure}

To show that StCE indeed addresses the problems posed by SC, we repeat our experiments in~\cref{sec:evidence-of-sc} by replacing \softmax with $\stablemax$. Our results, presented in \cref{fig:input_representations}, indeed show that $\stablemax$ leads to grokking in commonly studied settings \textit{without} regularization. Notably, this happens while the norm of the weights increases substantially (\cref{fig:input_representations}, middle). This suggests that while weight decay may lead to both grokking and a decreasing weight norm, the decreasing weight norm is not necessary for grokking. Overall, these results i) provide additional evidence for the importance of SC in preventing grokking, ii) suggest a novel activation function to address this problem, and iii) show that regularization or weight norm modification is not \textit{necessary} for grokking.

\begin{comment}
\begin{figure}
\centering
\begin{subfigure}[t]{.65\textwidth}
    \includegraphics[width=\linewidth]{grokking_iclr/figures/softermax_grokking.pdf}
\end{subfigure}%
\begin{subfigure}[t]{.35\textwidth}
    \raisebox{55pt}[0pt][0pt]{%
    \includegraphics[width=\linewidth]{grokking_iclr/figures/soft_exponential.pdf}}
    
\end{subfigure}%

\caption{Gokking with stable-cross-entropy loss and no regularization on three common grokking datasets using an MLP with 2 hidden layers of width 200. Regular CE  Loss with no regularization gets stuck at random level performance (which is 50\% for sparse parity) in these settings. 40\% of all pairs modulo 113 are used for training on the modular arithmetic tasks.}
\label{fig:stable_max}
\end{figure}
\end{comment}

\section{Diagnosing the Causes of Softmax Collapse}
\label{sec:nlm}

In the previous section we have shown that FP errors arise due to a combination of low losses and large logits, and shown that when FP errors are mitigated, grokking can be observed in conditions where it previously was not. In this section, we dive deeper and ask why extremely low losses and large logits appear in the first place in grokking tasks. We identify two main causes for this tendency: (i) easiness of overfitting in grokking tasks, and (ii) a training dynamic that sees gradients align with what we call \textit{naïve loss minimization} direction. After diagnosing the causes, the following section will use these insights to develop an optimization algorithm that avoids NLM in the first place.

\subsection{Ease of overfitting in grokking tasks}

The first important characteristic of grokking tasks that lead to SC is their ease of overfitting. It has been observed that as grokking datasets get larger, overfitting becomes harder, eventually leading to a regime where train and test performances increase in tandem \citep{power2022grokking, Nanda2023-hf, Varma2023}. It has also been shown that generalization can be delayed in the Sparse Parity task by increasing the amount of noise in the input, which makes overfitting easier \citep{Barak2022-el}. Here we investigate the opposite effect: that by decreasing the dimensionality of the input the data becomes harder to memorize, removing the delay in generalization.

To do this, we investigate the common grokking task of modular addition, but instead of the high-dimensional one-hot representations of the input integers, we use a more compact binary. More specifically, we assign each integer a distinct random binary vector of dimension 14.

Results confirm our hypothesis, showing that as input representations are decreased in dimension, overfitting is prevented and models generalize without need for regularization (\cref{fig:input_representations}, right). This also shows that modular addition only induces grokking depending on the choice of representation. These findings highlight the importance of understanding the training dynamics beyond the point of overfitting (i.e. point of achieving 100\% training accuracy), rather than focusing on the specifics of the modular arithmetic tasks as the key to explaining the delay in generalization.

\subsection{Naïve loss minimization}

We next identify a crucial training dynamic that commonly occurs in grokking tasks as a central cause for increasing logits and SC. We find that after reaching 100\% training accuracy, gradient updates are dominated by an update direction we term \textit{naïve loss minimization} (NLM). This direction does not change the model's decision boundary, but still decreases loss by simply scaling the logits of the predictions, in most cases through scaling of parameters (see below). This means that the logits will continue to increase until they inevitably lead to SC and zero terms in the training gradient. This stops the parameter updates in any direction, including NLM \textit{and} any other useful component that would have been included in the overall gradient. We now define NLM formally, and proceed to discuss why it might commonly be observed to deteriorate training in grokking tasks. Given the input $\mathbf{x} \in \mathcal{X}$, output $y \in \mathcal{Y}$, a predictor $f$ parametrized by $\params \in \mathbb{R}^m$ that outputs logits $\mathbf{z} = f(\params; \mathbf{x}) \in \mathbb{R}^{|\mathcal{Y}|}$, and a loss function $\mathcal{L}$, we now define Naïve Loss Minimization.

\begin{dfn}[Naïve Loss Minimization (\nlm)]
A function $d_\mathrm{NLM}: \mathbb{R}^m \to \mathbb{R}^m$  specifies a direction of naïve loss minimization if it decreases the loss,
\begin{equation}
    \mathcal{L}(f(\params + d_\mathrm{NLM}(\params);\cdot)) < \mathcal{L}(f(\params;\cdot)),
\end{equation}
while satisfying for some $c>1$:
\begin{equation}
    f(\params + d_\mathrm{NLM}(\params); \bm{x}) = c f(\params; \bm{x}), \quad \forall \mathbf{x} \in \mathcal{X},
\end{equation}

where $\mathcal{X}$ denotes the input space and $\mathcal{L}(f(\params + d_\mathrm{NLM}(\params);\cdot))$ is the total loss over the training dataset.
\end{dfn}

We find that under a large class of models, namely those that demonstrate \textit{positive homogeneity}, when training beyond 100\% training accuracy the direction of the weights is an NLM direction.
\begin{dfn}[Positive Homogeneity \citep{Lyu2019-sc}]
    A function $f$ is positively homogeneous of degree $L > 0$ if for all weights $\params$, inputs $\mathbf{x}$, and scalars $c > 0$, it satisfies:
\begin{equation}\label{eq:homogeneous}
f(c\params;\, \mathbf{x}) = c^{L} f(\params;\, \mathbf{x}) ~ .
\end{equation}

\begin{figure}[t]
\begin{subfigure}[t]{.33\textwidth}
    \includegraphics[width=\linewidth]{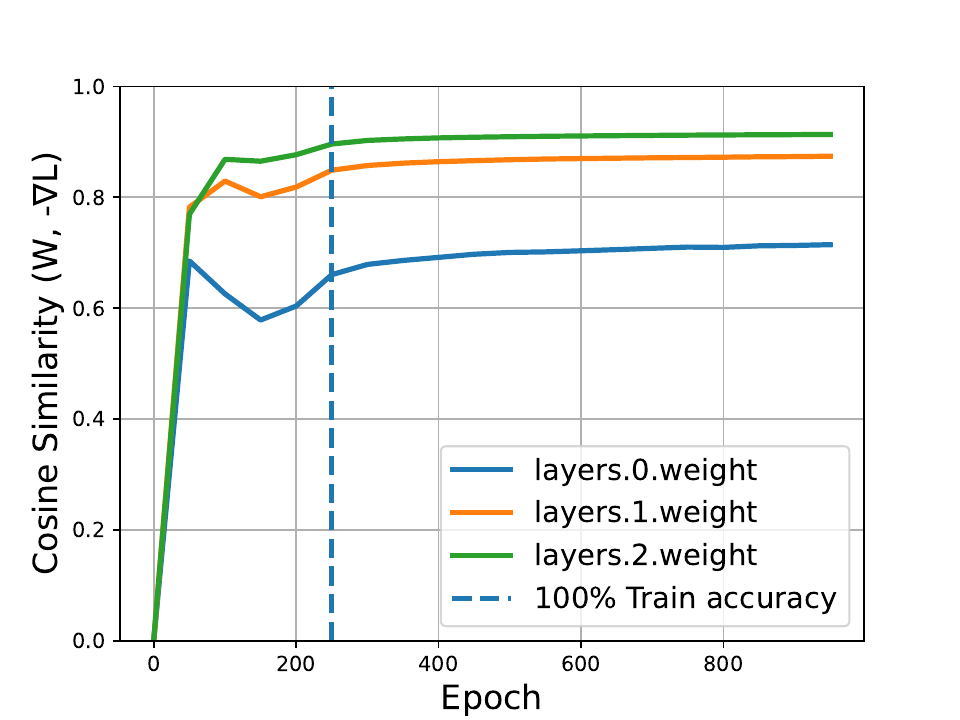}
    \caption{\vspace{-1mm}MLP without bias terms}
    \label{fig:nmm_no_bias}
\end{subfigure}%
\begin{subfigure}[t]{.33\textwidth}
    \includegraphics[width=\linewidth]{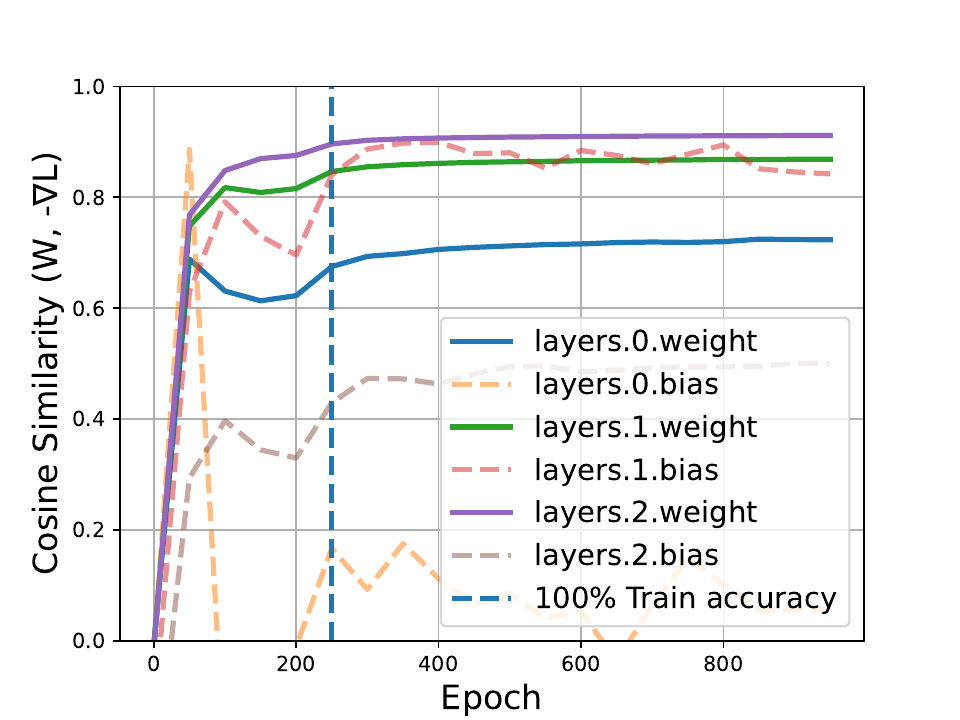}
    \caption{\vspace{-1mm}MLP with bias terms}
    \label{fig:nmm_bias}
\end{subfigure}
\begin{subfigure}[t]{.33\textwidth}
    \includegraphics[width=\linewidth]{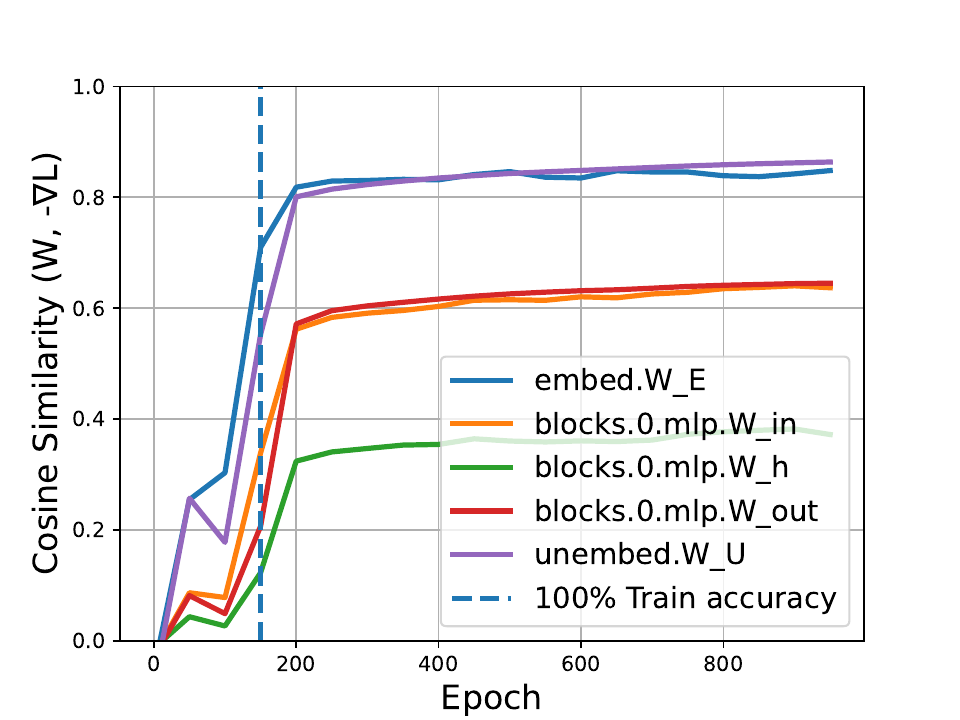}
    \caption{\vspace{-1mm}Transformer with bias terms}
    \label{fig:nmm_no_bias_transformer}
\end{subfigure}
\vspace{-1mm}
    \caption{MLPs with (\textbf{a}) and without (\textbf{b}) bias terms trained on modular addition receive updates that are significantly aligned with the direction of \nlm beyond the point of overfitting. In (\textbf{c}) we show these results for a selection of parameters for our one layer transformer. We highlight the embed and unembed matrices as well as the weights of the MLP. These are highlighted in the plot using the notation from \cite{elhage2021mathematical}.\vspace{-4mm}}
\label{fig:proof_of_nmm}
\end{figure}
When $f$ is a homogeneous neural network, $L$ corresponds to the number of layers.
\end{dfn}

In the case of homogeneous networks, training beyond 100\% training accuracy, scaling the logits always leads to a decrease in the training loss. Therefore, $d_\mathrm{NLM}(\params)= \alpha\params$ for $\alpha>0$ is an NLM direction, as it results in $f(\params + d_\mathrm{NLM}(\params); \bm{x}) = f((1+\alpha)\params; \bm{x}) = (1+\alpha)^Lf(\params; \bm{x})$, where the second equality follows from \cref{eq:homogeneous}. 

Many neural network architectures, such as ReLU MLPs and transformers without bias terms, are \emph{positively homogeneous} or \emph{approximately homogeneous} in the case of transformers \citep{homogeneous_transformers}. While more complex deep learning models with skip connections and bias terms are not homogeneous, they have been shown to be quasi-homogeneous~\citep{kunin2023asymmetricmaximummarginbias} and in most cases -- including all of the models in this work, the last layer is homogeneous. This means that for non-homogeneous models scaling the weights of the last layer corresponds to a direction of \nlm. 

The fact that the gradients converge to the direction of the weights has been studied in previous works \citep{NEURIPS2020_c76e4b2f, ji2018gradient, ji20182, Lyu2019-sc} to prove that homogeneous networks converge in direction under gradient flow and gradient descent (GD), and they perform normalized margin maximization even beyond the point of $100\%$ training accuracy \citep{Lyu2019-sc}. However, we argue that gradient alignment also results in scaling of the logits which can lead to SC and put an end to the margin maximization described in \cite{Lyu2019-sc}, when working with limited floating point precision. While we study delayed generalization, the link between training trajectories and generalization is already established in prior art~\citep{birdal2021intrinsic,andreeva2024topological}.

\paragraph{Evidence of naïve loss minimization}
In practice, we observe that in MLPs and transformers with and without bias terms, the gradients quickly become aligned with the direction of the weights after the point of overfitting (\cref{fig:proof_of_nmm}). Particularly for the later layers of the models, the cosine similarity between the parameter updates and the NLM direction goes up to 0.9 for the output layers. While models with bias terms are not homogeneous and there is no theoretical guarantee that scaling the weights will reduce the SCE loss, in practice, we observe very similar behavior in MLPs with (\cref{fig:nmm_bias}) and without (\cref{fig:nmm_no_bias}) bias terms. In the case of a one-layer transformer, the alignment is stronger for the embed and unembed matrices but also substantial for the MLP weights (\cref{fig:nmm_no_bias_transformer}).

\section{Mitigating Naïve Loss Minimization Leads to Grokking}
\label{sec:avoiding_nmm}

While we have shown in \cref{sec:floating_points} that avoiding numerical instabilities eventually leads to generalization, we can also target the \nlm  process that causes these numerical issues. To do this, we design an optimizer that only preserves the part of the gradient orthogonal to the direction of the weights.

\subsection{$\ograd$: An optimizer to prevent NLM}

\begin{figure*}[t]
\begin{subfigure}[t]{.32\textwidth}
    \includegraphics[width=\linewidth]{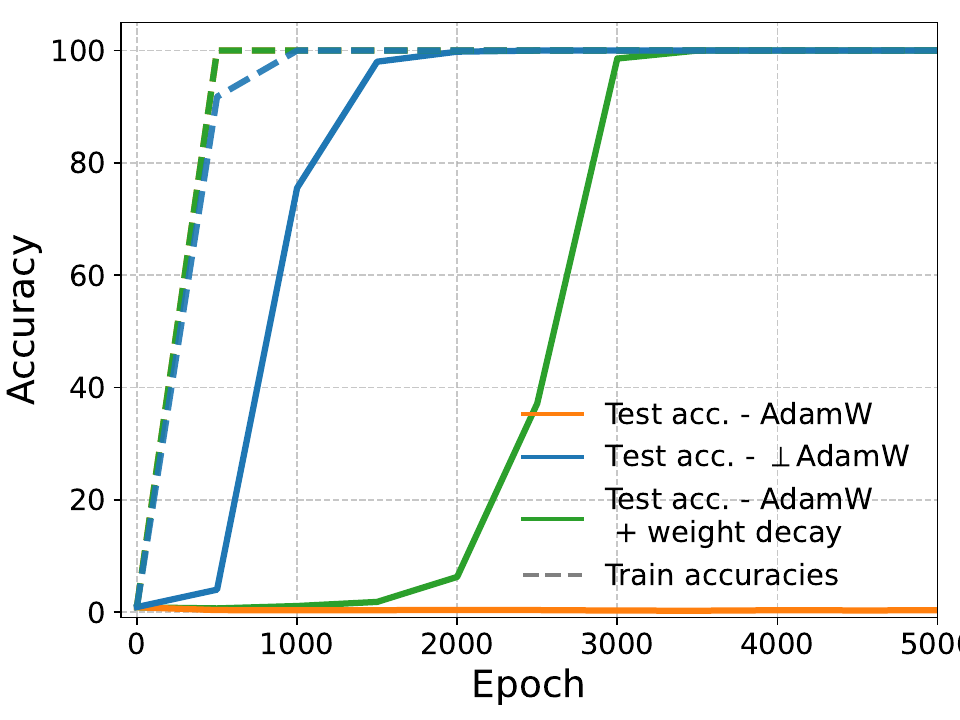}
    \caption{\vspace{-1.5mm}Transformer, subtract. mod 113}
    \label{fig:orthogonal_gradient}
\end{subfigure}
\hfill
\begin{subfigure}[t]{.32\textwidth}
    \includegraphics[width=\linewidth]{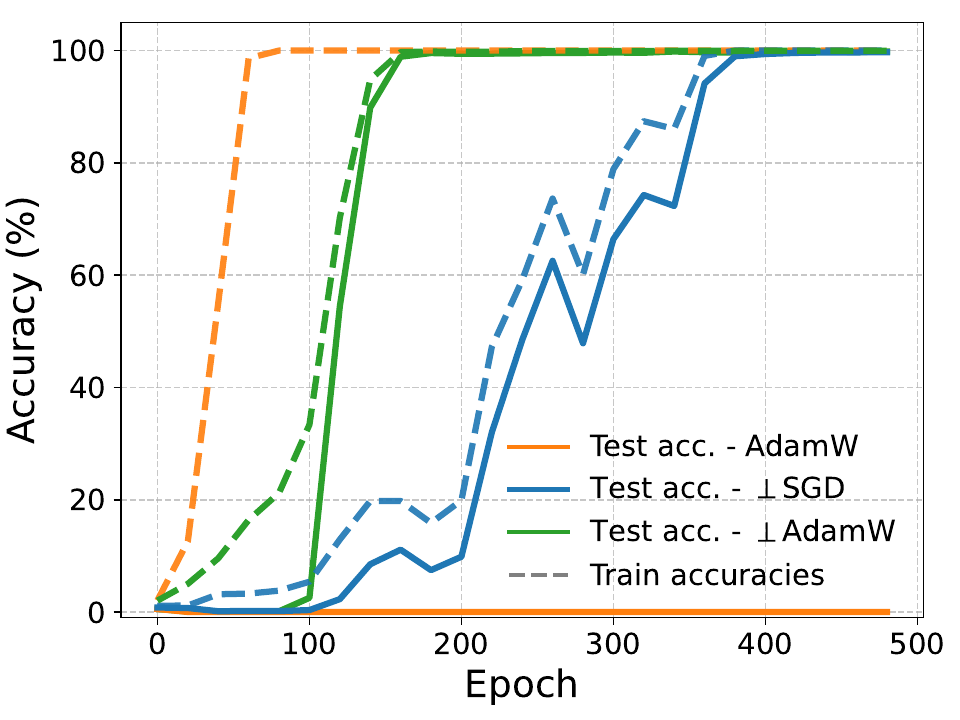}
    \caption{\vspace{-1.5mm}MLP, addition mod 113}
\label{fig:orthogonal_gradient_mlp}
\end{subfigure}
\hfill
\begin{subfigure}[t]{.32\textwidth}
  \centering
  \includegraphics[width=\linewidth]{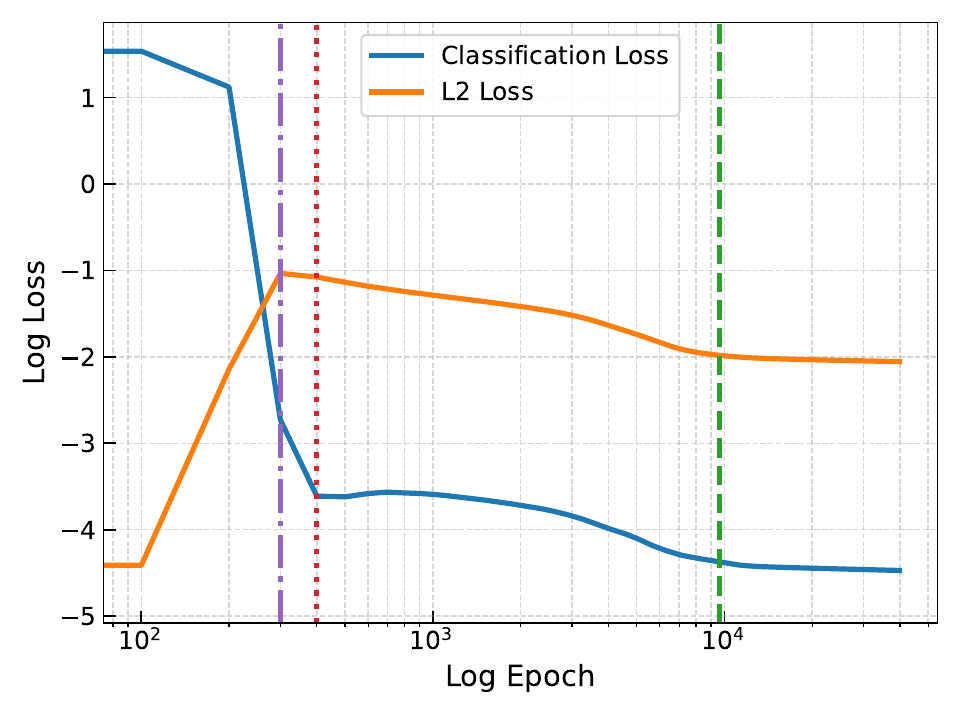}
  \caption{\vspace{-1.5mm}Trade-off between L2 and SCE}
  \label{fig:l2_vs_sce}
\end{subfigure}
\vspace{-1mm}
\caption{Comparing $\perp$AdamW and $\perp$SGD with baseline optimizers and AdamW with weight decay on (\textbf{a}) a transformer trained on subtraction mod 113 and  (\textbf{b}) an MLP trained on addition modulo 113. In (\textbf{c}) we highlight the trade-off between L2 regularization and SCE loss, initially SCE loss is reduced at the cost of increasing the L2 loss but eventually the two losses decrease simultaneously (\cref{sec:explain_existing_methods}). \vspace{-7mm}}
\end{figure*}

We propose a new optimizer, $\perp$\textbf{Grad} (read ``ortho-grad''), that updates the weights based only on the part of the gradient that is orthogonal to the current direction of the weights:

\begin{dfn}[$\perp$Grad]
    We propose the following update rule for a given iteration $t \in \mathbb{N}$: 
    \begin{equation}
        \bm{\theta}_{t+1} = \bm{\theta}_t - \eta \nabla_{\perp} \loss(\bm{\theta}_t),
    \end{equation}
    where the orthogonal component 
 of the gradient, $\nabla_{\perp} \loss(\bm{\theta}_t)$, is obtained by projection onto the hyperplane orthogonal to the current weight vector: 
 \begin{equation}
     \nabla_{\perp} \loss(\bm{\theta}_t) = \nabla \loss(\bm{\theta}_t) - \left( \frac{\bm{\theta}_t^\top \nabla \loss(\bm{\theta}_t)}{\bm{\theta}_t^\top \bm{\theta}_t} \right) \bm{\theta}_t.
 \end{equation}
%NMMGrad converges since $\nabla_{\perp} \loss(\bm{\theta}_t)$ is a descent direction.
\end{dfn}
\begin{prop}\label{prop:NLMGrad}
    Assuming $\nabla_{\perp} \loss(\bm{\theta}_t)\neq \mathbf{0}$, $\exists~\beta>0$ such
that for any learning rate $0 <\eta<\beta$, taking the step $\eta\nabla_{\perp} \loss(\bm{\theta}_t)$ reduces the loss. In other words, any nonzero $\nabla_{\perp} \loss(\bm{\theta}_t)$ is a descent direction.
\end{prop}
\vspace{-7pt}
\begin{proof}[Sketch of the proof.]
We show that any $\nabla_{\perp} \loss(\bm{\theta}_t) \in \mathbb{R}^m\backslash\{\mathbf{0}\}$ is a descent direction by demonstrating that $\left\langle -\nabla_{\perp} \loss(\bm{\theta}_t), \nabla\loss(\bm{\theta}_t) \right\rangle < 0$. For a full proof we refer the reader to~\cref{app:proofs}.
\end{proof}

This projection of the gradient can be incorporated into different optimizers. In \cref{fig:orthogonal_gradient}, we show results for $\perp$AdamW and $\perp$SGD, the $\perp$Grad versions of AdamW and SGD respectively. These results show that $\perp$Grad optimizers lead to generalization without a phase of initial overfitting, in contexts where no improvement in test performance is usually observed without weight decay. We note that similar projections of the gradients have been used in other settings to mitigate the effects of momentum in invariant layers \citep{adamp2020}, stabilize training \cite{wang2024achieving} or as one part in a more complex optimizer \citep{kosson2024rotational}. We design \ograd as a more precise intervention that directly prevents scaling along the NLM direction.

\begin{figure}[t]
\begin{subfigure}[t]{.48\textwidth}
    \centering
    \includegraphics[width=\linewidth]{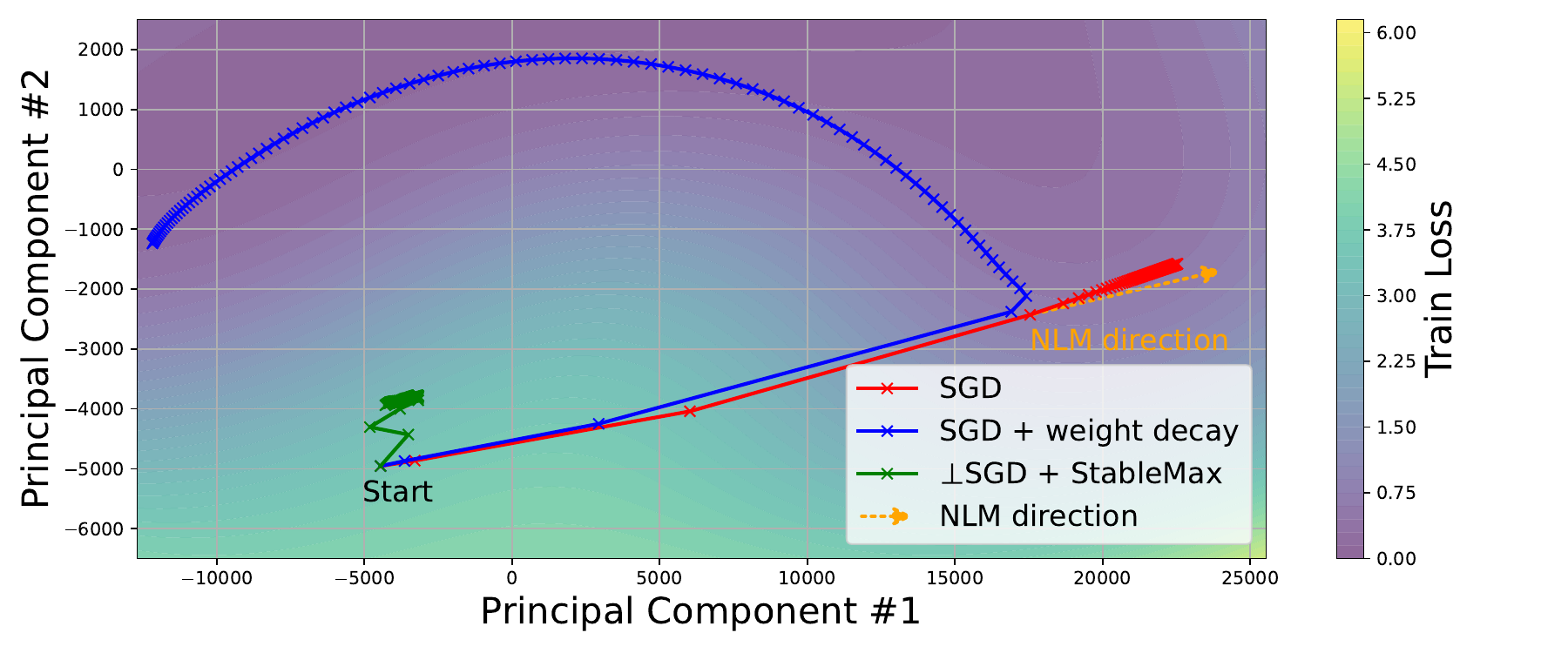}
    \caption{Training loss landscape}
    \label{fig:trajectory_plot}
\end{subfigure}
\hfill
\begin{subfigure}[t]{.48\textwidth}
  \centering
  \includegraphics[width=\linewidth]{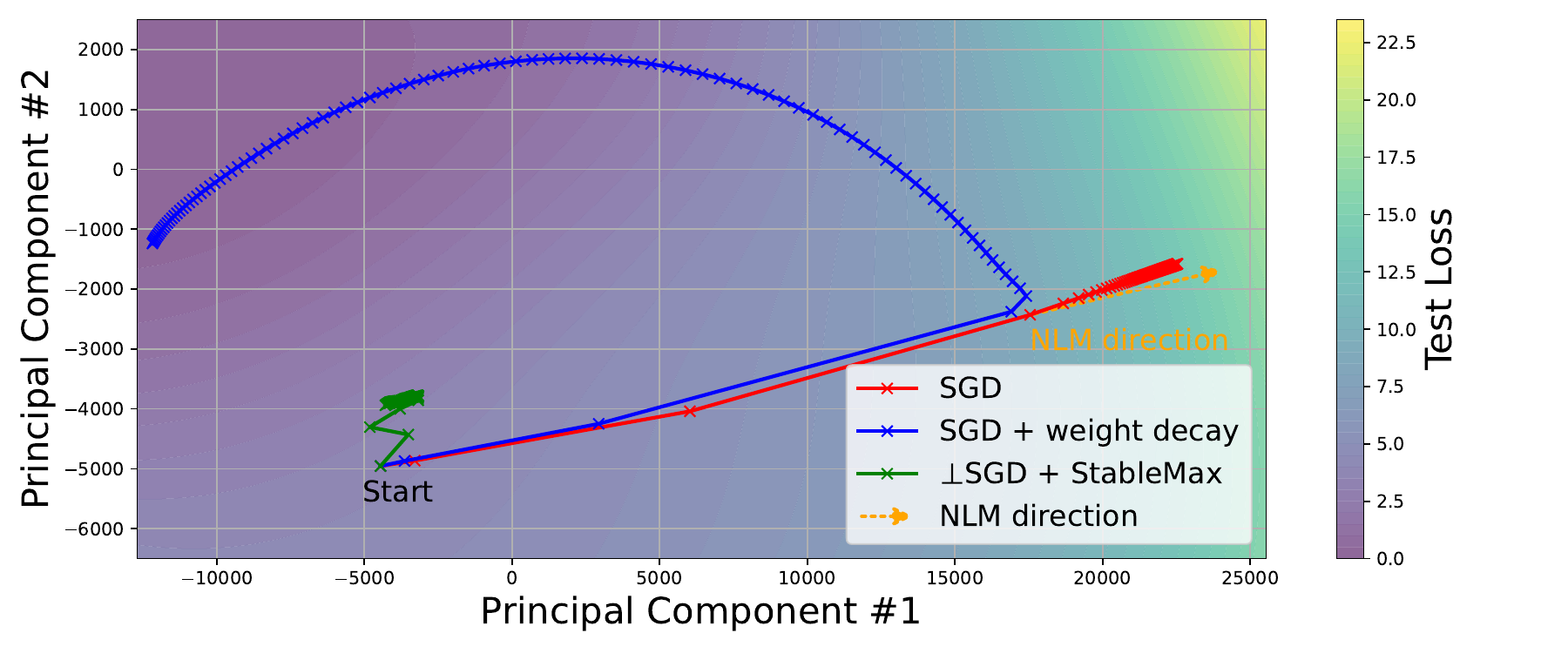}
  \caption{Test loss landscape}
\end{subfigure}
\vspace{-3mm}
\caption{Model trajectories in in parameter space projected to 2D over the SCE loss landscape. SGD with weight decay starts along the same trajectory as SGD decreasing the training loss \textbf{(a)} but increasing the test loss \textbf{(b)}.\vspace{-6 mm}}
\label{fig:loss_landscape}
\end{figure}

In~\cref{fig:loss_landscape}, we compare the trajectories of models using SGD with and without weight decay to our new $\perp$SGD optimizer. SGD models start on a similar trajectory, reducing the training loss but increasing the test loss, until the model with weight decay changes direction and starts minimizing both the train and test loss. In contrast, the model using $\perp$SGD moves directly in a direction that minimizes both the train and test loss. While SGD with weight decay eventually reaches a point of lower loss, note that $\perp$SGD reaches 100\% test accuracy within 400 iterations (\cref{fig:orthogonal_gradient}). Beyond showing how $\perp$SGD prevents NLM, \cref{fig:loss_landscape} also suggests that weight decay induces grokking by avoiding NLM. In the following, we highlight that the success of several methods to induce grokking can be explained from this perspective.

% Eventually, the test loss starts to decrease for the model with weight decay while SGD without weight decay continues along NLM direction. The NLM direction is highlighted as the direction that corresponds to scaling up all the weights by 20\% for the first model to reach 100\% train accuracy with SGD. Note that the "detour" taken by the model with weight decay corresponds to the delay in generalization observed during grokking and that $\perp$SGD decreases the test loss from the beginning, without a delay in generalization
\subsection{Explaining the success of existing methods for grokking}\label{sec:explain_existing_methods}

In light of our findings, we are able to explain the success of several previously proposed methods to induce grokking. We find that these methods also lead to grokking by mitigating NLM and avoiding the FP errors that come with extremely low losses.

\paragraph{Weight decay} We have argued that the problem faced in grokking is that the ease of overfitting leads to \nlm, which corresponds to scaling up the weights for homogeneous networks. Since weight decay corresponds to pulling back the weights along this same direction at every step during training, it is unsurprising, given our findings, that it is the most reliable way to induce grokking. 

To explain why generalization tends to be delayed when using weight decay, as opposed to $\perp$Grad, we look at it from the perspective of L2 regularization which is equivalent to weight decay for SGD. In~\cref{fig:l2_vs_sce}, we see an initial phase where classification loss decreases, at the cost of the L2 loss.  Eventually, the decrease in classification loss from \nlm stops outweighing the increase in L2 loss, meaning that only updates that are not aligned with the NLM direction are followed. This explains why weight decay leads to generalization in grokking tasks but only after scaling along the \nlm direction no longer decreases the overall loss. This balance between weight decay and classification loss is similar to the rotational equilibrium studied in \cite{kosson2024rotational}.

We argue that the main roles of weight decay are preventing floating point errors and preventing NLM. This is in line with recent findings about the role of weight decay in deep learning \citep{andriushchenko2023needweightdecaymodern} which point to the fact that it increases the effective learning rate and avoids floating point issues when using mixed-precision training in LLMs.

\paragraph{MSE loss on shallow networks}
While cross-entropy loss can be reduced indefinitely by scaling the logits through \nlm, this is not the case with MSE loss. When using MSE loss the logits can overshoot the target, meaning that larger logits often do not lead to a lower MSE loss. This explains why \cite{Barak2022-el}, \cite{Kumar2023-hz}, and \cite{Lyu2023-ga} observed grokking with MSE loss without regularization. Interestingly, networks with more than one hidden layer do not generalize in these same settings (\cref{fig:alpha_parameter}).

\paragraph{Delaying generalization by scaling the weights}
While the lazy training dynamics described in \cite{Kumar2023-hz} explain an important part of why scaling the weights delays generalization, we show that the reason that regularization is often needed to exit this lazy training regime is that scaling the weights or the logits facilitates SC.  In \cref{app:scaling_weights}, we show that the setting used in \cite{liu2023grokking} to induce grokking on MNIST with SCE also induces SC which prevents further learning in the absence of weight decay.
\section{Related Work}
\paragraph{Grokking}
\cite{power2022grokking} introduced grokking and showed that weight decay can consistently induce it in algorithmic tasks. \cite{Nanda2023-hf} were able to reverse engineer the inner workings of a grokked transformer and found progress measures for grokking induced by weight decay. \cite{chughtai2023toy} generalized the findings from \cite{Nanda2023-hf} and showed grokked networks use group representations to solve group composition tasks, although some of these findings were disputed in \cite{stander2023grokking} which propose that grokked networks learn a coset based algorithm for these same tasks. \citet{mallinar2024emergence} has shown that grokking is not specific to neural networks or gradient-based optimization and cannot be predicted from the training or test loss. \cite{Varma2023} argued that grokking is driven by weight decay favoring more efficient solutions and \cite{liu2023grokking} hypothesized that the weight norm of the models needs to be in a ``Goldilock's zone'' to generalize. \cite{Kumar2023-hz} and \cite{Lyu2023-ga} connected grokking to a transition between ``lazy training'' \citep{Chizat_Oyallon_Bach_2018} and feature learning, and \cite{Kumar2023-hz} showed that this can happen without regularization in the case of shallow networks with MSE loss. Grokking has also been described as a phase transition by \cite{vzunkovivc2024grokking}, \cite{Lyu2023-ga} and \cite{rubin2024grokking}. \cite{humayun2024deep} show that in many settings, neural networks undergo grokking-like transitions in their adversarial robustness. This aligns with the findings of \cite{Lyu2019-sc} which attributed this increased robustness to a bias of SGD towards a max-margin solution which was proven for homogeneous models. \cite{beck2024grokkingedgelinearseparability} also connected grokking to the linear separability of the training data.

\paragraph{Numerical instability in deep learning} 
Numerical instability is a common issue in deep learning \cite{stability}, especially when dealing with mixed precision training \cite{andriushchenko2023needweightdecaymodern}. It is known that the \softmax function is particularly prone to numerical stability problems although this often comes in the form of overflow in the exponential \citep{stability} and not from absorption errors in the sum as observed in this case. In the grokking setting, \cite{Nanda2023-hf} showed that the slingshots observed in \cite{slingshot-mechanism} can be explained by a very similar mechanism to the one involved in SC, although \cite{Nanda2023-hf} do not use it to explain any grokking phenomena beyond these spikes that sometimes appear in the training process in grokking tasks. We believe the slingshots observed in \cite{slingshot-mechanism} could be a mechanism to prevent full SC, explaining why slingshots can lead to grokking without weight decay in some settings. This is further discussed in \cref{app:slingshots}. Issues with numerical instability when training beyond overfitting with increasing learning rates were also observed in \cite{Lyu2019-sc}.

\begin{comment}

Emergent abilities~\citep{huang2024unified}
\citet{power2022grokking} 
modular addition~\cite{} and multiplication~\cite{doshi2024grokking}
\cite{wang2024grokked} find that transformers can learn implicit reasoning, but only through grokking.
\cite{fan2024deep} reveal that deep neural networks can be more susceptible to grokking than its shallower counterparts. This is supported by \cite{humayun2024deep} arguing that deep networks always grok. These findings make the extent of our study even more interesting for practical settings. 
phase transition from lazy to rich regime~\citep{kumar2023grokking,mohamadi2024you,Lyu2023-ga,rubin2023droplets}, learning local rules~\cite{vzunkovivc2024grokking}.
phase transition and emergence~\citep{clauw2024information}, formation of geometric arrangement of circuits in the input space~\cite{humayun2024grokking}
Group theoretic~\cite{stander2023grokking,chughtai2023toy}

compression and robustness~\cite{liu2023grokking,tan2023understanding}: weight norm (metric) of the neural network is actually a
sufficient condition for grokking

\end{comment}

\vspace{-2mm}
\section{Conclusion and Discussion}
\vspace{-2mm}
In this work, we show that naïve loss minimization (NLM) and floating point errors can explain why generalization is delayed in grokking and why it often does not happen without regularization. Using this insight, we are able to explain the success of existing methods to induce grokking. Motivated by our findings, we further design a simple modification to the \softmax that induces grokking by avoiding floating point errors and an optimizer that avoids the delay in generalization in grokking by preventing NLM. 

\paragraph{Limitations \& future work} While this work explains several surprising aspects of grokking settings, several questions remain. Notably, we focus our study of NLM on homogeneous or approximately homogeneous models. A a formal characterization  quasi-homogenous models could shed light on this kind of dynamics for models including skip connections and bias terms. Additionally, our explanation for why weight decay causes grokking could be enhanced by an analysis of its impact on the effective learning rate as a potential explanation for the sudden nature of grokking.

\footnotesize
\paragraph{Acknowledgments}
This work was supported by the UKRI Centre for Doctoral Training in Safe and Trusted AI
[EP/S0233356/1]. TB acknowledges support from the Engineering and Physical Sciences Research Council [grant EP/X011364/1].
TB was supported by a UKRI Future Leaders Fellowship [grant number MR/Y018818/1]. 

\bibliography{iclr2025_conference}

\begin{thebibliography}{41}
\providecommand{\natexlab}[1]{#1}
\providecommand{\url}[1]{\texttt{#1}}
\expandafter\ifx\csname urlstyle\endcsname\relax
  \providecommand{\doi}[1]{doi: #1}\else
  \providecommand{\doi}{doi: \begingroup \urlstyle{rm}\Url}\fi

\bibitem[Andreeva et~al.(2024)Andreeva, Dupuis, Sarkar, Birdal, and
  {\c{S}}im{\c{s}}ekli]{andreeva2024topological}
Rayna Andreeva, Benjamin Dupuis, Rik Sarkar, Tolga Birdal, and Umut
  {\c{S}}im{\c{s}}ekli.
\newblock Topological generalization bounds for discrete-time stochastic
  optimization algorithms.
\newblock In \emph{Adv. Neural Inf. Process. Syst.}, 2024.

\bibitem[Barak et~al.(2022)Barak, Edelman, Goel, Kakade, Malach, and
  Zhang]{Barak2022-el}
Boaz Barak, Benjamin Edelman, Surbhi Goel, Sham Kakade, Eran Malach, and Cyril
  Zhang.
\newblock Hidden progress in deep learning: Sgd learns parities near the
  computational limit.
\newblock \emph{Advances in Neural Information Processing Systems}, 35, 2022.

\bibitem[Beck et~al.(2024)Beck, Levi, and
  Bar-Sinai]{beck2024grokkingedgelinearseparability}
Alon Beck, Noam Levi, and Yohai Bar-Sinai.
\newblock Grokking at the edge of linear separability.
\newblock \emph{arXiv preprint arXiv:2410.04489}, 2024.

\bibitem[Birdal et~al.(2021)Birdal, Lou, Guibas, and
  Simsekli]{birdal2021intrinsic}
Tolga Birdal, Aaron Lou, Leonidas~J Guibas, and Umut Simsekli.
\newblock Intrinsic dimension, persistent homology and generalization in neural
  networks.
\newblock \emph{Advances in Neural Information Processing Systems}, 34, 2021.

\bibitem[Chizat et~al.(2018)Chizat, Oyallon, and
  Bach]{Chizat_Oyallon_Bach_2018}
Lénaïc Chizat, Edouard Oyallon, and F.~Bach.
\newblock On lazy training in differentiable programming.
\newblock \emph{Advances in neural information processing systems}, pp.\
  2933–2943, December 2018.
\newblock ISSN 1049-5258.

\bibitem[Chughtai et~al.(2023)Chughtai, Chan, and Nanda]{chughtai2023toy}
Bilal Chughtai, Lawrence Chan, and Neel Nanda.
\newblock A toy model of universality: Reverse engineering how networks learn
  group operations.
\newblock In \emph{International Conference on Machine Learning}. PMLR, 2023.

\bibitem[D'Angelo et~al.(2023)D'Angelo, Andriushchenko, Varre, and
  Flammarion]{andriushchenko2023needweightdecaymodern}
Francesco D'Angelo, Maksym Andriushchenko, Aditya Varre, and Nicolas
  Flammarion.
\newblock Why do we need weight decay in modern deep learning?
\newblock \emph{arXiv preprint arXiv:2310.04415}, 2023.

\bibitem[Deng et~al.(2018)Deng, Birdal, and Ilic]{deng2018ppfnet}
Haowen Deng, Tolga Birdal, and Slobodan Ilic.
\newblock Ppfnet: Global context aware local features for robust 3d point
  matching.
\newblock In \emph{Proceedings of the IEEE conference on computer vision and
  pattern recognition}, 2018.

\bibitem[Deng(2012)]{deng2012mnist}
Li~Deng.
\newblock The mnist database of handwritten digit images for machine learning
  research.
\newblock \emph{IEEE Signal Processing Magazine}, 29\penalty0 (6):\penalty0
  141--142, 2012.

\bibitem[Devlin et~al.(2019)Devlin, Chang, Lee, and Toutanova]{devlin2018bert}
Jacob Devlin, Ming-Wei Chang, Kenton Lee, and Kristina Toutanova.
\newblock {BERT}: Pre-training of deep bidirectional transformers for language
  understanding.
\newblock In Jill Burstein, Christy Doran, and Thamar Solorio (eds.),
  \emph{Proceedings of the Conference of the North {A}merican Chapter of the
  Association for Computational Linguistics: Human Language Technologies}, pp.\
   4171--4186. Association for Computational Linguistics, June 2019.

\bibitem[Dugas et~al.(2000)Dugas, Bengio, B\'{e}lisle, Nadeau, and
  Garcia]{softplus}
Charles Dugas, Yoshua Bengio, Fran\c{c}ois B\'{e}lisle, Claude Nadeau, and
  Ren\'{e} Garcia.
\newblock Incorporating second-order functional knowledge for better option
  pricing.
\newblock In T.~Leen, T.~Dietterich, and V.~Tresp (eds.), \emph{Advances in
  Neural Information Processing Systems}, volume~13. MIT Press, 2000.

\bibitem[Elhage et~al.(2021)Elhage, Nanda, Olsson, Henighan, Joseph, Mann,
  Askell, Bai, Chen, Conerly, DasSarma, Drain, Ganguli, Hatfield-Dodds,
  Hernandez, Jones, Kernion, Lovitt, Ndousse, Amodei, Brown, Clark, Kaplan,
  McCandlish, and Olah]{elhage2021mathematical}
Nelson Elhage, Neel Nanda, Catherine Olsson, Tom Henighan, Nicholas Joseph, Ben
  Mann, Amanda Askell, Yuntao Bai, Anna Chen, Tom Conerly, Nova DasSarma, Dawn
  Drain, Deep Ganguli, Zac Hatfield-Dodds, Danny Hernandez, Andy Jones, Jackson
  Kernion, Liane Lovitt, Kamal Ndousse, Dario Amodei, Tom Brown, Jack Clark,
  Jared Kaplan, Sam McCandlish, and Chris Olah.
\newblock A mathematical framework for transformer circuits.
\newblock \emph{Transformer Circuits Thread}, 2021.
\newblock https://transformer-circuits.pub/2021/framework/index.html.

\bibitem[Gromov(2023)]{Gromov2023-nh}
Andrey Gromov.
\newblock Grokking modular arithmetic.
\newblock \emph{arXiv preprint arXiv:2301.02679}, 2023.

\bibitem[Heo et~al.(2021)Heo, Chun, Oh, Han, Yun, Kim, Uh, and Ha]{adamp2020}
Byeongho Heo, Sanghyuk Chun, Seong~Joon Oh, Dongyoon Han, Sangdoo Yun, Gyuwan
  Kim, Youngjung Uh, and Jung-Woo Ha.
\newblock Adamp: Slowing down the slowdown for momentum optimizers on
  scale-invariant weights.
\newblock In \emph{International Conference on Learning Representations}, 2021.

\bibitem[Humayun et~al.(2024)Humayun, Balestriero, and
  Baraniuk]{humayun2024deep}
Ahmed~Imtiaz Humayun, Randall Balestriero, and Richard Baraniuk.
\newblock Deep networks always grok and here is why.
\newblock \emph{arXiv preprint arXiv:2402.15555}, 2024.

\bibitem[Ji \& Telgarsky(2018)Ji and Telgarsky]{ji20182}
Ziwei Ji and Matus Telgarsky.
\newblock Risk and parameter convergence of logistic regression.
\newblock \emph{arXiv preprint arXiv:1803.07300}, 2018.

\bibitem[Ji \& Telgarsky(2019)Ji and Telgarsky]{ji2018gradient}
Ziwei Ji and Matus Telgarsky.
\newblock Gradient descent aligns the layers of deep linear networks.
\newblock In \emph{7th International Conference on Learning Representations,
  ICLR}, 2019.

\bibitem[Ji \& Telgarsky(2020)Ji and Telgarsky]{NEURIPS2020_c76e4b2f}
Ziwei Ji and Matus Telgarsky.
\newblock Directional convergence and alignment in deep learning.
\newblock In H.~Larochelle, M.~Ranzato, R.~Hadsell, M.F. Balcan, and H.~Lin
  (eds.), \emph{Advances in Neural Information Processing Systems}, volume~33,
  pp.\  17176--17186. Curran Associates, Inc., 2020.

\bibitem[Kloberdanz et~al.(2022)Kloberdanz, Kloberdanz, and Le]{stability}
Eliska Kloberdanz, Kyle~G Kloberdanz, and Wei Le.
\newblock Deepstability: A study of unstable numerical methods and their
  solutions in deep learning.
\newblock In \emph{Proceedings of the 44th International Conference on Software
  Engineering}, pp.\  586--597, 2022.

\bibitem[Kosson et~al.(2024)Kosson, Messmer, and Jaggi]{kosson2024rotational}
Atli Kosson, Bettina Messmer, and Martin Jaggi.
\newblock Rotational equilibrium: How weight decay balances learning across
  neural networks, 2024.

\bibitem[Krizhevsky et~al.(2012)Krizhevsky, Sutskever, and
  Hinton]{krizhevsky2012imagenet}
Alex Krizhevsky, Ilya Sutskever, and Geoffrey~E Hinton.
\newblock Imagenet classification with deep convolutional neural networks.
\newblock In \emph{Advances in neural information processing systems},
  volume~25, pp.\  1097--1105, 2012.

\bibitem[Kumar et~al.(2024)Kumar, Bordelon, Gershman, and
  Pehlevan]{Kumar2023-hz}
Tanishq Kumar, Blake Bordelon, Samuel~J. Gershman, and Cengiz Pehlevan.
\newblock Grokking as the transition from lazy to rich training dynamics.
\newblock In \emph{The Twelfth International Conference on Learning
  Representations}, 2024.

\bibitem[Kunin et~al.(2023)Kunin, Yamamura, Ma, and
  Ganguli]{kunin2023asymmetricmaximummarginbias}
Daniel Kunin, Atsushi Yamamura, Chao Ma, and Surya Ganguli.
\newblock The asymmetric maximum margin bias of quasi-homogeneous neural
  networks.
\newblock In \emph{The Eleventh International Conference on Learning
  Representations}, 2023.

\bibitem[Liu et~al.(2023{\natexlab{a}})Liu, Michaud, and
  Tegmark]{liu2023omnigrok}
Ziming Liu, Eric~J Michaud, and Max Tegmark.
\newblock Omnigrok: Grokking beyond algorithmic data.
\newblock In \emph{The Eleventh International Conference on Learning
  Representations}, 2023{\natexlab{a}}.

\bibitem[Liu et~al.(2023{\natexlab{b}})Liu, Zhong, and
  Tegmark]{liu2023grokking}
Ziming Liu, Ziqian Zhong, and Max Tegmark.
\newblock Grokking as simplification: A nonlinear complexity perspective.
\newblock In \emph{UniReps: the First Workshop on Unifying Representations in
  Neural Models}, 2023{\natexlab{b}}.

\bibitem[Lv et~al.(2024)Lv, Xie, Sun, Kang, and Yan]{lv2024language}
Ang Lv, Ruobing Xie, Xingwu Sun, Zhanhui Kang, and Rui Yan.
\newblock Language models" grok" to copy.
\newblock \emph{arXiv preprint arXiv:2409.09281}, 2024.

\bibitem[Lyu \& Li(2020)Lyu and Li]{Lyu2019-sc}
Kaifeng Lyu and Jian Li.
\newblock Gradient descent maximizes the margin of homogeneous neural networks.
\newblock In \emph{International Conference on Learning Representations}, 2020.

\bibitem[Lyu et~al.(2024)Lyu, Jin, Li, Du, Lee, and Hu]{Lyu2023-ga}
Kaifeng Lyu, Jikai Jin, Zhiyuan Li, Simon~Shaolei Du, Jason~D. Lee, and Wei Hu.
\newblock Dichotomy of early and late phase implicit biases can provably induce
  grokking.
\newblock In \emph{The Twelfth International Conference on Learning
  Representations}, 2024.

\bibitem[Mallinar et~al.(2024)Mallinar, Beaglehole, Zhu, Radhakrishnan, Pandit,
  and Belkin]{mallinar2024emergence}
Neil Mallinar, Daniel Beaglehole, Libin Zhu, Adityanarayanan Radhakrishnan,
  Parthe Pandit, and Mikhail Belkin.
\newblock Emergence in non-neural models: grokking modular arithmetic via
  average gradient outer product.
\newblock \emph{arXiv preprint arXiv:2407.20199}, 2024.

\bibitem[Merrill et~al.(2020)Merrill, Ramanujan, Goldberg, Schwartz, and
  Smith]{homogeneous_transformers}
William Merrill, Vivek Ramanujan, Yoav Goldberg, Roy Schwartz, and Noah~A.
  Smith.
\newblock Parameter norm growth during training of transformers.
\newblock \emph{CoRR}, abs/2010.09697, 2020.

\bibitem[Nanda et~al.(2023)Nanda, Chan, Lieberum, Smith, and
  Steinhardt]{Nanda2023-hf}
Neel Nanda, Lawrence Chan, Tom Lieberum, Jess Smith, and Jacob Steinhardt.
\newblock Progress measures for grokking via mechanistic interpretability.
\newblock In \emph{The Eleventh International Conference on Learning
  Representations}, 2023.

\bibitem[Power et~al.(2022)Power, Burda, Edwards, Babuschkin, and
  Misra]{power2022grokking}
Alethea Power, Yuri Burda, Harri Edwards, Igor Babuschkin, and Vedant Misra.
\newblock Grokking: Generalization beyond overfitting on small algorithmic
  datasets.
\newblock \emph{arXiv preprint arXiv:2201.02177}, 2022.

\bibitem[Qi et~al.(2017)Qi, Su, Mo, and Guibas]{qi2017pointnet}
Charles~R Qi, Hao Su, Kaichun Mo, and Leonidas~J Guibas.
\newblock Pointnet: Deep learning on point sets for 3d classification and
  segmentation.
\newblock In \emph{Proceedings of the IEEE conference on computer vision and
  pattern recognition}, pp.\  652--660, 2017.

\bibitem[Radford et~al.(2019)Radford, Wu, Child, Luan, Amodei, and
  Sutskever]{radford2019language}
Alec Radford, Jeff Wu, Rewon Child, David Luan, Dario Amodei, and Ilya
  Sutskever.
\newblock Language models are unsupervised multitask learners.
\newblock 2019.

\bibitem[Rubin et~al.(2024)Rubin, Seroussi, and Ringel]{rubin2024grokking}
Noa Rubin, Inbar Seroussi, and Zohar Ringel.
\newblock Grokking as a first order phase transition in two layer networks.
\newblock In \emph{The Twelfth International Conference on Learning
  Representations}, 2024.

\bibitem[Russakovsky et~al.(2015)Russakovsky, Deng, Su, Krause, Satheesh, Ma,
  Huang, Karpathy, Khosla, Bernstein, Berg, and Fei-Fei]{ILSVRC15}
Olga Russakovsky, Jia Deng, Hao Su, Jonathan Krause, Sanjeev Satheesh, Sean Ma,
  Zhiheng Huang, Andrej Karpathy, Aditya Khosla, Michael Bernstein,
  Alexander~C. Berg, and Li~Fei-Fei.
\newblock {ImageNet Large Scale Visual Recognition Challenge}.
\newblock \emph{International Journal of Computer Vision (IJCV)}, 115\penalty0
  (3):\penalty0 211--252, 2015.
\newblock \doi{10.1007/s11263-015-0816-y}.

\bibitem[Stander et~al.(2024)Stander, Yu, Fan, and
  Biderman]{stander2023grokking}
Dashiell Stander, Qinan Yu, Honglu Fan, and Stella Biderman.
\newblock Grokking group multiplication with cosets.
\newblock In \emph{Forty-first International Conference on Machine Learning},
  2024.

\bibitem[Thilak et~al.(2022)Thilak, Littwin, Zhai, Saremi, Paiss, and
  Susskind]{slingshot-mechanism}
Vimal Thilak, Etai Littwin, Shuangfei Zhai, Omid Saremi, Roni Paiss, and Josh
  Susskind.
\newblock The slingshot mechanism: An empirical study of adaptive optimizers
  and the grokking phenomenon.
\newblock In \emph{NeurIPS Workshop}, 2022.

\bibitem[Varma et~al.(2023)Varma, Shah, Kenton, Kram{\'a}r, and
  Kumar]{Varma2023}
Vikrant Varma, Rohin Shah, Zachary Kenton, J{\'a}nos Kram{\'a}r, and Ramana
  Kumar.
\newblock Explaining grokking through circuit efficiency.
\newblock \emph{arXiv preprint arXiv:2309.02390}, 2023.

\bibitem[Wang et~al.(2024)Wang, Min, and Wu]{wang2024achieving}
Mingze Wang, Zeping Min, and Lei Wu.
\newblock Achieving margin maximization exponentially fast via progressive norm
  rescaling.
\newblock \emph{arXiv preprint arXiv:2311.14387}, 2024.

\bibitem[{\v{Z}}unkovi{\v{c}} \& Ilievski(2024){\v{Z}}unkovi{\v{c}} and
  Ilievski]{vzunkovivc2024grokking}
Bojan {\v{Z}}unkovi{\v{c}} and Enej Ilievski.
\newblock Grokking phase transitions in learning local rules with gradient
  descent.
\newblock \emph{Journal of Machine Learning Research}, 25\penalty0
  (199):\penalty0 1--52, 2024.

\end{thebibliography}
\bibliographystyle{iclr2025_conference}

\newpage
\appendix
\section*{Appendix}
In support of the main paper,~\cref{app:proofs} presents the proofs for the propositions in the paper,~\cref{app:additional_findings} includes additional findings that support our main results, and~\cref{app:further_discussion} provides further discussion on conditions that lead to grokking.
\section{Proofs}\label{app:proofs}
\begin{proof}[Proof of~\cref{prop:stablemax}]
\begin{align}
    \softmax\left(g\left(x_i\right)\right) &= \frac{e^{g(x_i)}}{\sum_j e^{g(x_j)}}\\
    &= \begin{cases}
\frac{e^{\log(x_i+1)}}{\sum_j e^{\log(x_j+1)}} & \text{if } x_i \geq 0, \\
\frac{e^{-\log(-x_i+1)}}{\sum_j e^{-\log(-x_j+1)}} & \text{if } x_i < 0
\end{cases}\\
&= \begin{cases}
\frac{x_i+1}{\sum_j x_j+1} & \text{if } x_i \geq 0, \\
\frac{\frac{1}{-x_i+1}}{\sum_j \frac{1}{-x_j+1}} & \text{if } x_i < 0
\end{cases}\\
&= \stablemax(x_i).
\end{align}
\end{proof}

\begin{proof}[Proof of~\cref{prop:NLMGrad}]
To prove that any nonzero $-\nabla_{\perp} \loss(\bm{\theta}_t)$ is a descent direction, we need to show that $\left\langle -\nabla_{\perp} \loss(\bm{\theta}_t), \nabla\loss(\bm{\theta}_t) \right\rangle < 0$, assuming $\nabla_{\perp} \loss(\bm{\theta}_t) \neq \mathbf{0}$:
    \begin{equation}
        \left\langle \nabla\loss(\bm{\theta}_t), -\nabla\loss(\bm{\theta}_t) + \left( \frac{\bm{\theta}_t^\top \nabla \loss(\bm{\theta}_t)}{\bm{\theta}_t^\top \bm{\theta}_t} \right)\bm{\theta}_t  \right\rangle \leq 0.
    \end{equation}
    Expanding this yields:
    \begin{align}
        -\left\|\nabla\loss(\bm{\theta}_t)\right\|^2_2 + 
        \left\langle  \nabla\loss(\bm{\theta}_t), \bm{\theta}_t \frac{\bm{\theta}_t^\top \nabla \loss(\bm{\theta}_t)}{\bm{\theta}_t^\top \bm{\theta}_t}\right\rangle
        &\leq 0.
    \end{align}
    Since the inequality is unaffected by the scaling of the left hand side, we can, without loss of generality, assume that the gradients are normalized, leading to:
    \begin{align}\label{eq:incidence}
        \left\langle \nabla\loss(\bm{\theta}_t), \bm{\theta}_t \frac{\bm{\theta}_t^\top \nabla \loss(\bm{\theta}_t)}{\bm{\theta}_t^\top \bm{\theta}_t}\right\rangle
        &{\leq} 1.
    \end{align}
    Since $\bm{\theta}_t \frac{\bm{\theta}_t^\top \nabla \loss(\bm{\theta}_t)}{\bm{\theta}_t^\top \bm{\theta}_t}$ denotes the projection of the gradient onto the space spanned by the weights, $\langle\cdot,\cdot\rangle$ will measure the acute angle of incidence and hence~\cref{eq:incidence} holds, with equality iff $\nabla_{\perp} \loss(\bm{\theta}_t) = \mathbf{0}$, which is prevented by assumption. This proves that $-\nabla_{\perp} \loss(\bm{\theta}_t)$ is a descent direction while being perpendicular to the weights. %, the angle between $\loss(\bm{\theta}_t)$ and this projection will be the acute .
\end{proof}
We note that the \ograd stops when $\nabla_{\perp}\loss(\bm{\theta}_t) = \mathbf{0}$. If $\nabla\loss(\bm{\theta}_t) \neq \mathbf{0}$, this corresponds to the condition where the gradient is in the same direction with the parameter vector. $\nabla_{\perp}\loss(\bm{\theta}_t) = \mathbf{0}$ can also be the case if $\nabla\loss(\bm{\theta}_t) = \mathbf{0}$, which corresponds to the loss function being at a local optimum.

\section{Additional Findings}\label{app:additional_findings}

\subsection{Further evidence that SC prevents grokking} \label{app:sc_intervention}
\begin{wrapfigure}[12]{R}{0.38\textwidth}
            \vspace{-0.4cm}
            \begin{center}
    \includegraphics[width=\linewidth]{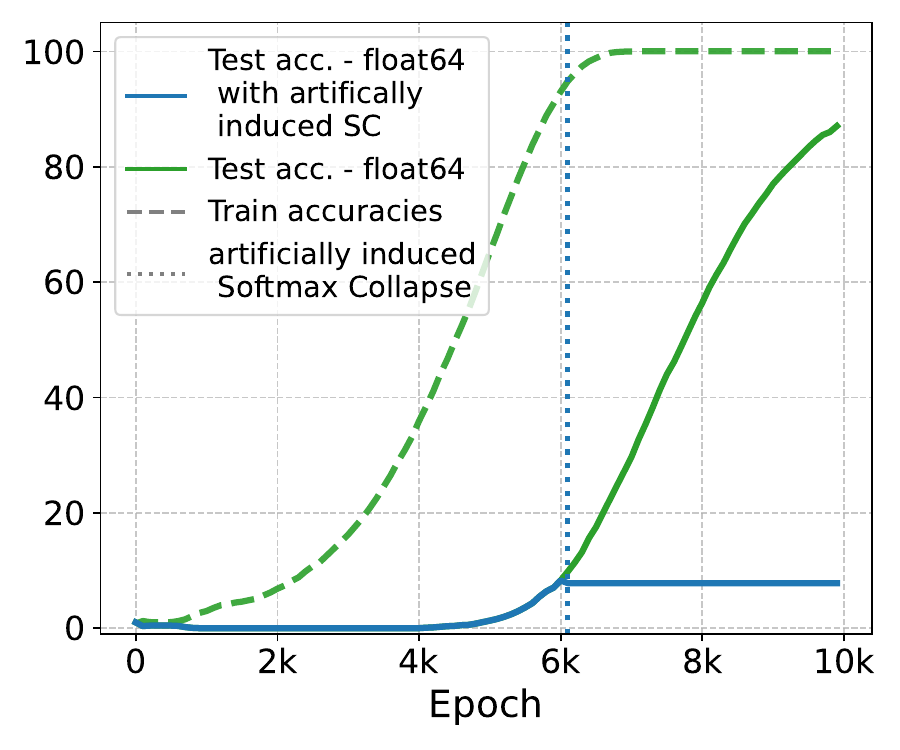}
            \end{center}
            \vspace{-12pt}
            \caption{Taking a model that would normally generalize (green) and artificially inducing SC has a very similar effect to the one observed in \cref{fig:grokking_stops}.\vspace{4mm}}
    \label{fig:sc_intervention}
\end{wrapfigure} 
While SC leads the gradient from correctly predicted samples to be zero, it does not do this for the incorrect classes. To validate that setting the gradients from the correct classes to zero is enough to stop learning, we do this artificially for a model that is generalizing and show that learning stops after this intervention. In \cref{fig:sc_intervention} we see that the baseline model shown in green generalizes, but this is stopped at epoch 6000 for the model shown in blue, after we perform this intervention.

The intervention is implemented by multiplying the logits for the right classes by 0 at each step after epoch 6000.

\begin{comment}
\begin{figure}[h]
    \centering
    \includegraphics[width=0.5\linewidth]{grokking_iclr_arxiv/figures/sc_intervention.pdf}
    \caption{Taking a model that would normally generalize (green) and artificially inducing SC has a very similar effect to the one observed in \cref{fig:grokking_stops}}.
    \label{fig:sc_intervention}
\end{figure}    
\end{comment}

\subsection{SGD with learning rate scheduling}
\begin{figure}[t]
\centering
\begin{subfigure}{.32\textwidth}
  \centering
  \includegraphics[width=\linewidth]{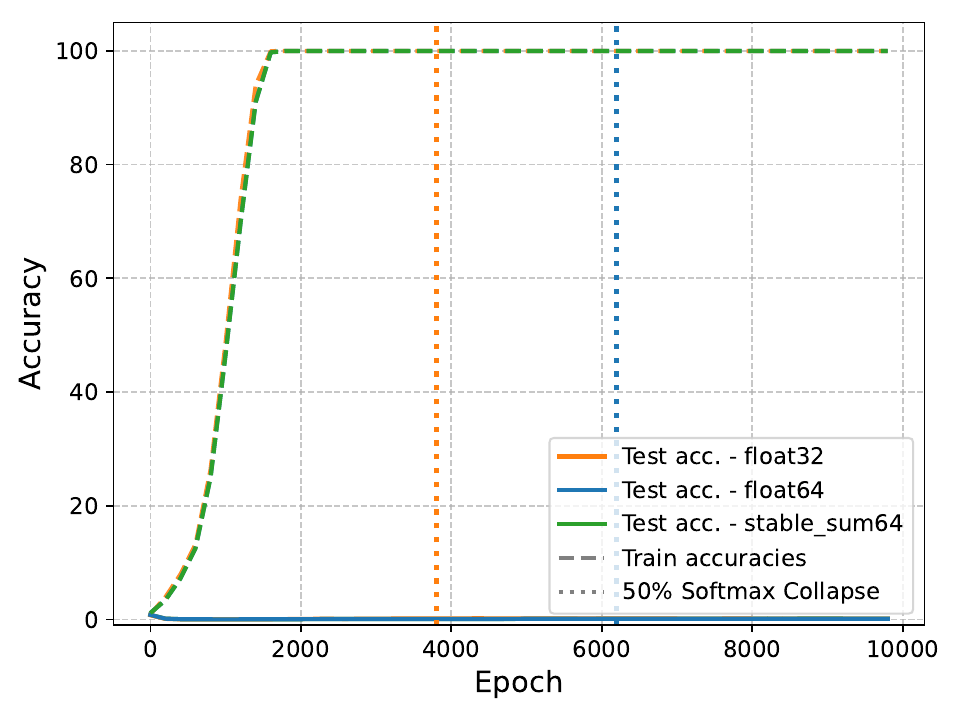}
  \caption{40\% training data}
  \label{fig:grokking_stops_40_lr_sch}
\end{subfigure}
\hfill
\begin{subfigure}{.32\textwidth}
  \centering
  \includegraphics[width=\linewidth]{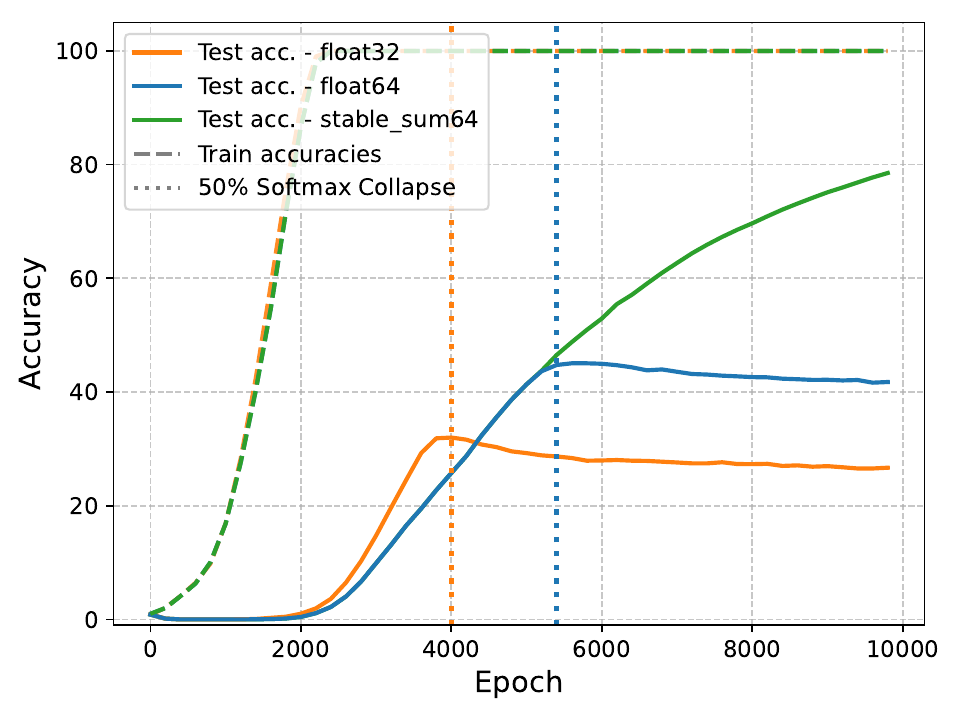}
  \caption{60\% training data}
  \label{fig:grokking_stops_60_lr_sch}
\end{subfigure}
\hfill
\begin{subfigure}{.32\textwidth}
  \centering
  \includegraphics[width=\linewidth]{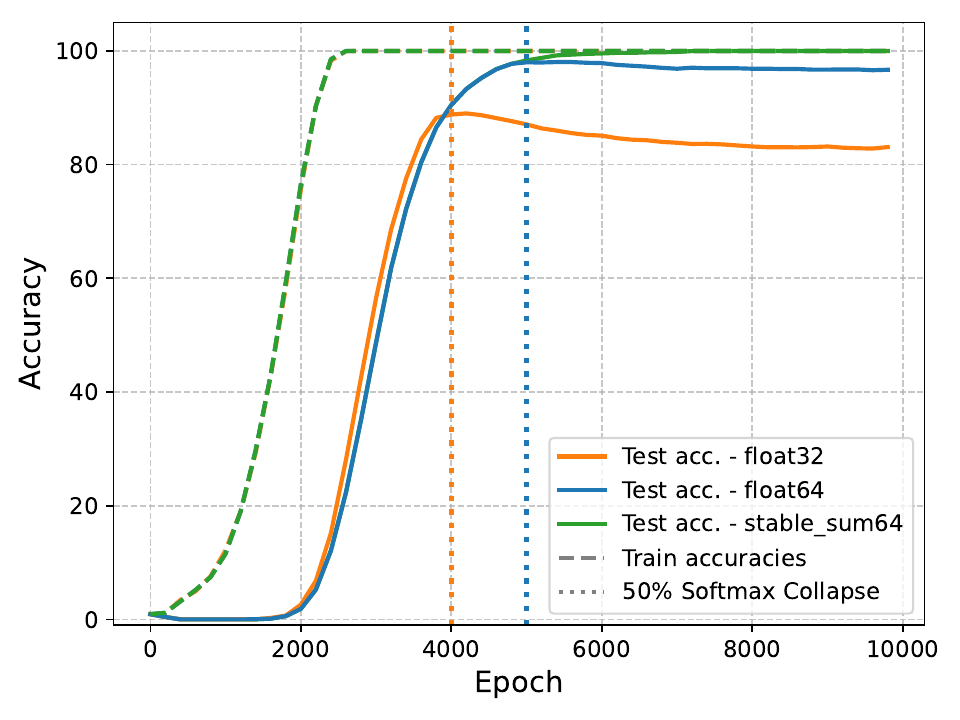}
  \caption{70\% training data}
\end{subfigure}
\caption{We show that the same dynamics observed in \cref{fig:grokking_stops} can be observed with a learning rate scheduler instead of AdamW. This shows that this is not due to an implicit bias of adaptive optimizers.}
\label{fig:grokking_stops_lr_sch}
\vspace{-5mm}
\end{figure}

To show that our results are not due to the inductive bias of adaptive moments in optimizers like AdamW, we replicate some of the AdamW results using SGD with a learning rate scheduler. Our scheduler is similar to the one in \cite{Lyu2019-sc} except at each step we divide the learning rate by the norm of the full gradient, instead of the loss. In \cref{fig:grokking_stops_lr_sch} we observe that SC also puts an end to grokking in this setting.
\vspace{5.75mm}\\

\vspace{-5mm}
\section{Effective Learning Rate}
\begin{wrapfigure}[13]{R}{0.4\textwidth}
            \vspace{-1.2cm}
            \begin{center}
    \includegraphics[width=0.4\textwidth]{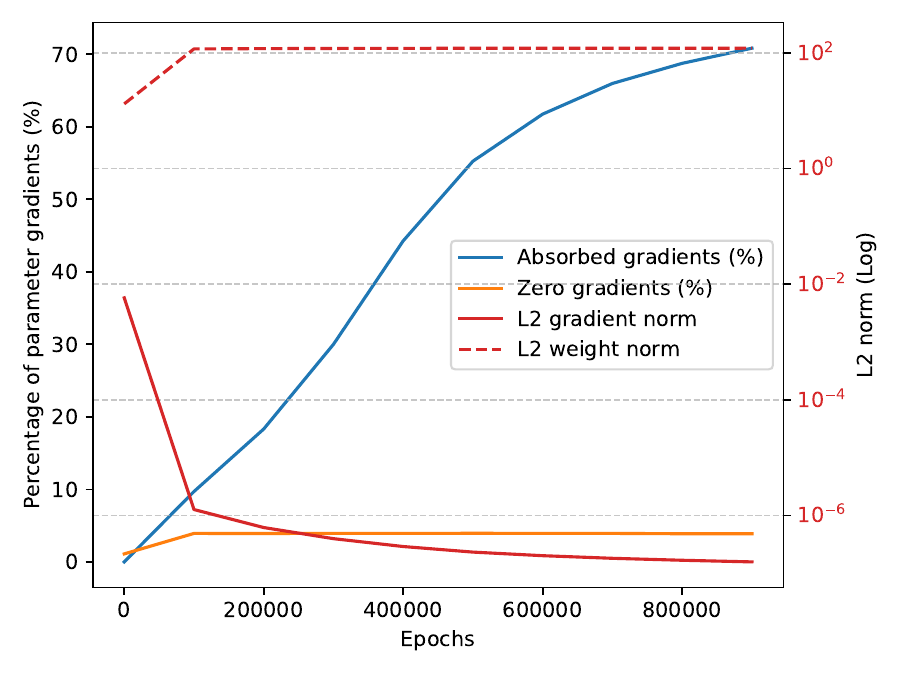}
            \end{center}
            \vspace{-10pt}
            \caption{Gradient absorption errors during training on addition modulo 113.}
            \label{fig:gradient_absorption}
\end{wrapfigure} 
Unexplored in the main paper, NLM also has the effect of reducing the effective learning rate. For a gradient update using regular gradient descent $\bm{\theta}_{t+1} = \bm{\theta}_t - \eta \nabla \loss(\bm{\theta}_t)$ it is easy to see that $||\bm{\theta}_{t+1} - \bm{\theta}_{t}|| \to 0$ as $||\nabla \loss(\bm{\theta}_t)|| \to 0$. This problem has been observed before when training beyond the point of overfitting, for example, \cite{Lyu2019-sc} addressed it by using a loss based learning rate scheduler to keep up with the gradient. Theoretically, an alternative could be to simply extend the duration of training. According to our hypothesis, training for long enough should eventually lead to generalization on grokking tasks if we prevent SC. However, we find that another kind of floating point error can also appears in these settings, namely, gradient absorption errors in the weights. 

For a weight $w$, gradient absorption errors happen when a gradient update is small enough that it leaves the weight unchanged. Using the notation outlined in this paper this can be formalized as $w -\eta \frac{\partial \mathcal{L}}{\partial w} \doteq w$. In \cref{fig:gradient_absorption} we show that this happens for an MLP trained with SGD on modular addition using 30\% of the training data. As the norm of the gradient decreases, the percentage of the gradients that are absorbed by the weights increases substantially. Note that the number of gradients that are \textit{exactly} zero remains stable while the number of absorbed gradients increases substantially.

\begin{comment}
\begin{figure}[h]
    \centering
    \includegraphics[width=0.5\linewidth]{grokking_iclr_arxiv/figures/gradient_norms.pdf}
    \caption{Gradient absorption errors during training on addition modulo 113.}
    \label{fig:gradient_absorption}
\end{figure}
\end{comment}
This issue is naturally mitigated by second order moments for adaptive optimizers like Adam and AdamW which is why they do not frequently appear. However, they do prevent us from showing grokking with vanilla gradient descent without any learning rate scheduling.

\subsection{Additional ways to induce grokking}
Beyond the interventions described in the main text, we highlight two additional ways to induce grokking that validate our hypothesis. 

\begin{figure}[t]
\centering
\begin{subfigure}{.33\textwidth}
  \centering
  \includegraphics[width=\linewidth]{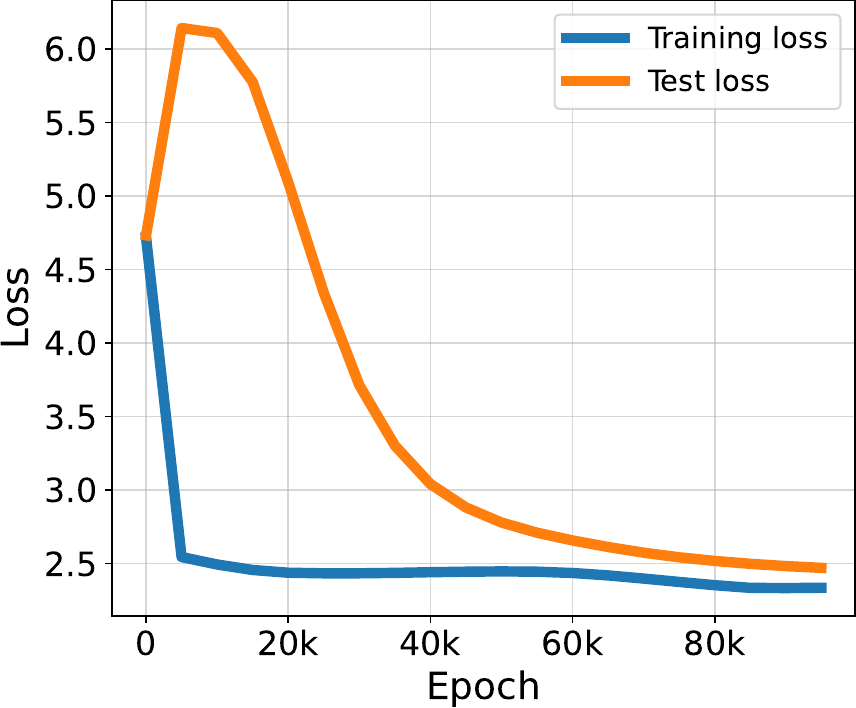}
  \caption{$\stablemax$}
  \label{fig:additional_grokking_stablemax}
\end{subfigure}%
\hfill
\begin{subfigure}{.33\textwidth}
  \centering
  \includegraphics[width=\linewidth]{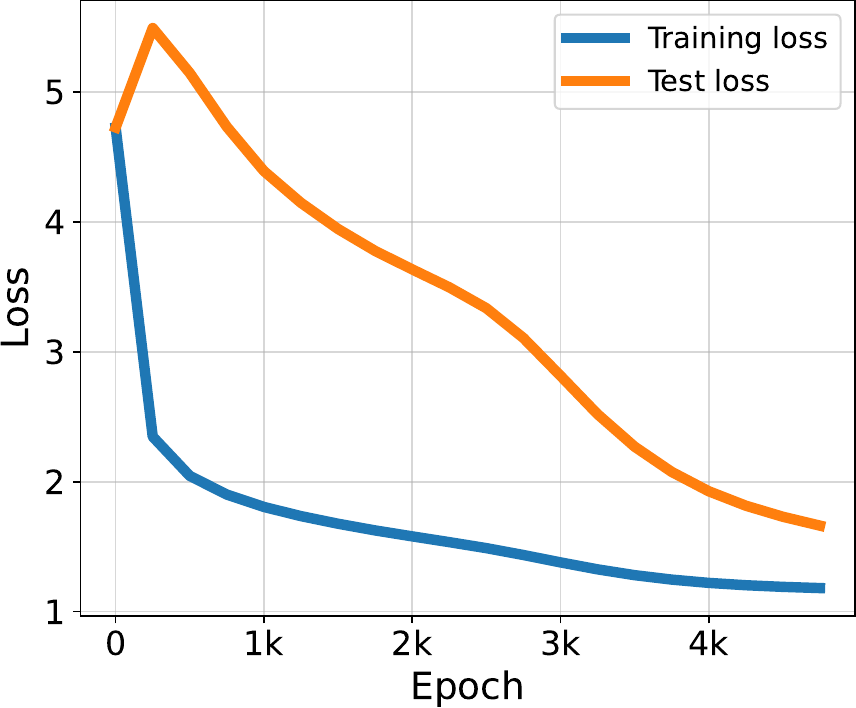}
  \caption{Logit regularization}
  \label{fig:additional_grokking_logit}
  
\end{subfigure}%
\hfill
\begin{subfigure}{.33\textwidth}
  \centering
  \includegraphics[width=\linewidth]{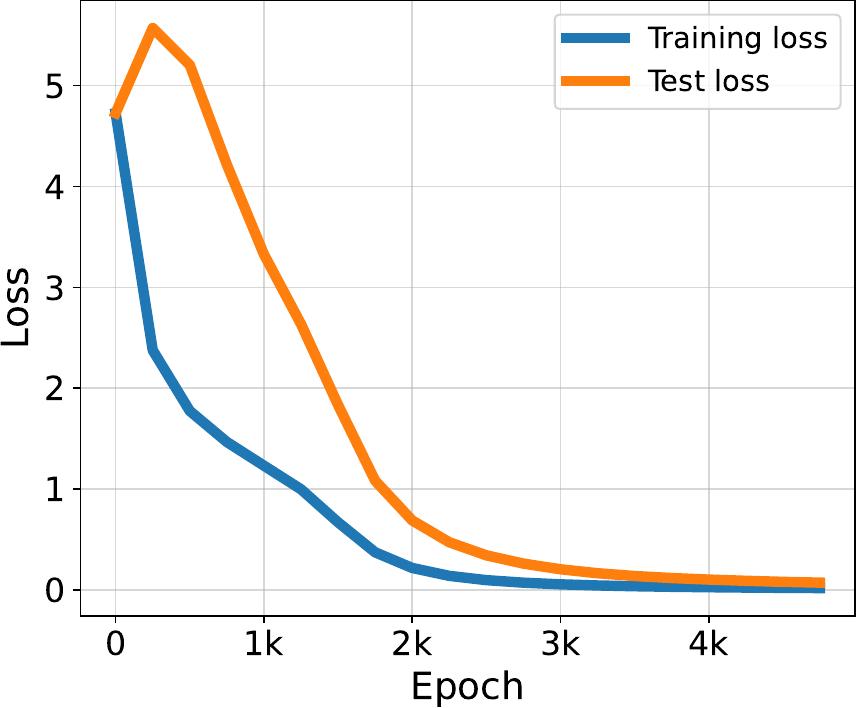}
  \caption{\tsoftmax}
  \label{fig:additional_grokking_taylor}
\end{subfigure}
\vspace{-3mm}
\caption{Train and test losses during grokking induced by three different interventions.\vspace{-2mm}}
\label{fig:additional_grokking}
\end{figure}

\paragraph{Logit norm regularization}
Since we argue that uncontrolled scaling of the logits is responsible for delaying grokking and leading to SC, we validate that preventing this scaling of the logits by adding the norm of the logits to the loss, leads to grokking without additional regularization (\cref{fig:additional_grokking_logit}).

\begin{wrapfigure}[20]{R}{0.4\textwidth}
\vspace{-0.5cm}
    \centering
    \includegraphics[width=\linewidth]{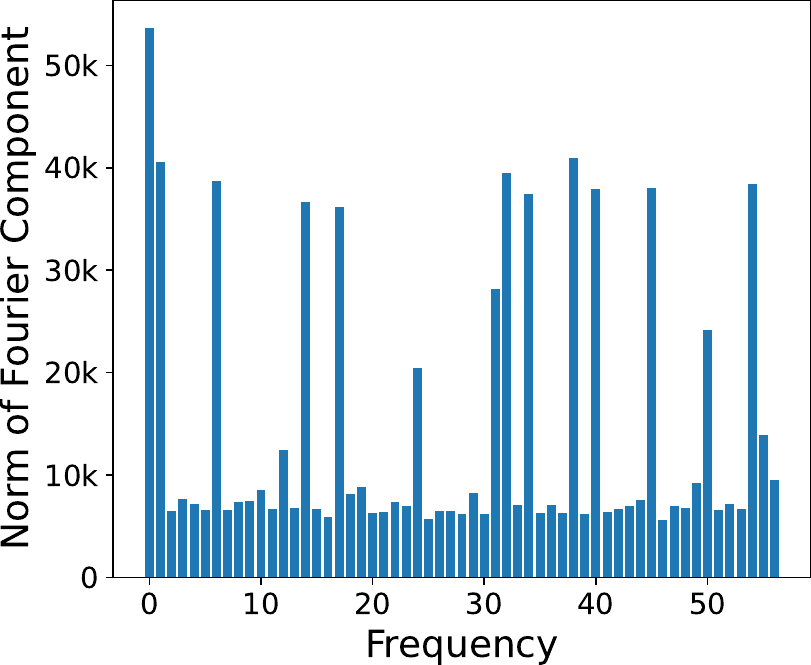}
    \vspace{-6mm}
    \caption{Fourier components of the weights of the output layer of an MLP trained on addition mod 113. Grokking is induced via $\stablemax$ and without weight decay.}
    \label{fig:fourier}
\end{wrapfigure}
\noindent\textbf{Taylor approximation of the Softmax.}
We have introduced $\stablemax$ as a change to the \softmax that leads to grokking without regularization. The motivation behind this is to prevent values in the sum of the \softmax that are very large or very close to zero. To this end, replacing the exponential with any function that is sub-exponential beyond a certain point should have a similar effect. To demonstrate, we perform a further experiment using the second order Taylor approximation of the exponential 
\begin{equation}
e^x\approx 1 + x + \frac{x^2}{2!},    
\end{equation}
replacing the $\exp$ in the \softmax. Since the Taylor approximation is decreasing for $x<0$, we subtract the minimum logit to avoid this part of the function.  We deem this version \tsoftmax.
In \cref{fig:additional_grokking} we see results similar to the ones in \cref{fig:stablemax_grokking} but showing the losses instead of the accuracies as well as results for two additional methods to induce grokking. 
Note that our implementation of \tsoftmax (\cref{fig:additional_grokking_taylor}) introduces an additional implicit regularization similar to the one in  \cref{fig:additional_grokking_logit}, due to the gradient flowing through the subtraction of the mean. % also introduces an additional regularization effect similar to the one 
While this effectively combines the effects of \cref{fig:additional_grokking_stablemax} and \cref{fig:additional_grokking_logit}, leading to grokking faster than the other two methods, our main paper shows results using $\stablemax$ as a cleaner intervention that does not introduce this additional regularization effect. 

% above \tolga{what is the 'one above'?}. This is why we show our main results using $\stablemax$ as a cleaner intervention that does not introduce this additional regularization effect. 

%Note that our implementation of \tsoftmax (\cref{fig:additional_grokking_taylor}) introduces an implicit regularization similar to the one in  \cref{fig:additional_grokking_logit}, effectively combining the effects of \cref{fig:additional_grokking_stablemax} and \cref{fig:additional_grokking_logit}, explaining why it leads to grokking faster than the other two methods. 

\subsection{Solution Learned During Grokking  Without Weight Decay}\label{app:fourier}

\begin{comment}
    \begin{wrapfigure}[20]{R}{0.4\textwidth}
            \vspace{-0.5cm}
            \begin{center}
    \includegraphics[width=0.4\textwidth]{grokking_iclr_arxiv/figures/fourier_components.pdf}
            \end{center}
            \vspace{-10pt}
            \caption{Fourier components of the weights of the output layer of an MLP trained on addition mod 113. Grokking is induced with $\stablemax$ and without weight decay. \tolga{this figure is terribly cropped.}}
            \label{fig:fourier}
        \end{wrapfigure}
\end{comment}

Weight decay has been identified as potentially responsible for inducing the periodic structures in the weights studied in \cite{Nanda2023-hf}. In \cref{fig:fourier} we show that MLPs that grok without weight decay on modular addition show a similar sparsity in Fourier space as the one observed in \cite{Nanda2023-hf}. While these are very superficial results, they suggest that these structures can emerge without a weight decay--induced ``clean up'' phase as described in \cite{Nanda2023-hf}.

\section{Further Discussion on Conditions that Lead to Grokking}\label{app:further_discussion}
\subsection{L1 regularization and grokking}\label{app:l1_grokking}

%\paragraph{L1 regularization}
While it has been observed that L1 regularization can lead to grokking in some settings, \cite{Nanda2023-hf}
consistently found no grokking with L1 regularization and transformers and this setting has received substantially less attention than weight decay. 

We observe that \nlm scales the weights along their current direction. This means that larger weights are scaled more than small weights. However, while the sign of the gradient from L1 regularization depends on the sign of the weights, the magnitude of this gradient does not depend on the magnitude of the weights. This means that, particularly on deep networks or transformers with with large weights, L1 can sometimes be insufficient to prevent \nlm and the subsequent SC. 
%In Appendix \ref{app:l1_grokking} we show two examples, one where L1 regularization avoids SC and leads to grokking, and one where SC happens despite L1 regularization and grokking does not happen.

\subsection{Delaying generalization by scaling the weights} \label{app:scaling_weights}

\begin{figure}[t]
\centering
\begin{subfigure}{.33\textwidth}
  \centering
  \includegraphics[width=\linewidth]{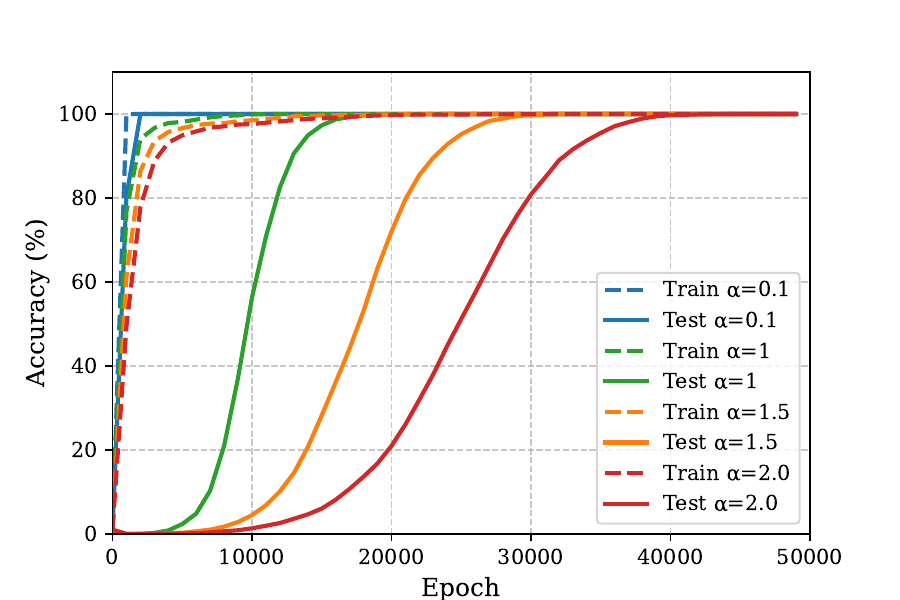}
  \caption{MSE: 1 hidden layer}
\end{subfigure}%
\hfill
\begin{subfigure}{.33\textwidth}
  \centering
  \includegraphics[width=\linewidth]{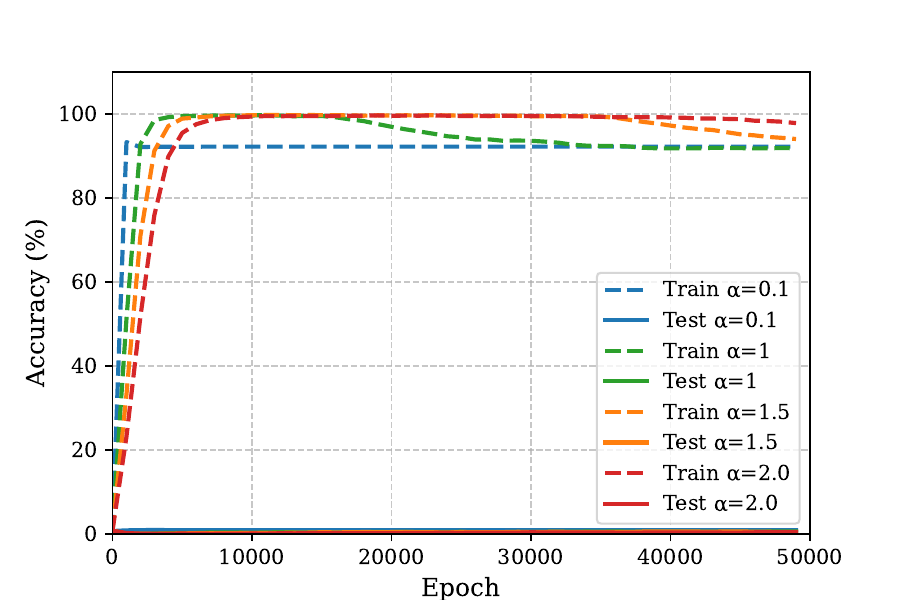}
  \caption{MSE: 2 hidden layers}
\end{subfigure}%
\hfill
\begin{subfigure}{.33\textwidth}
  \centering
  \includegraphics[width=\linewidth]{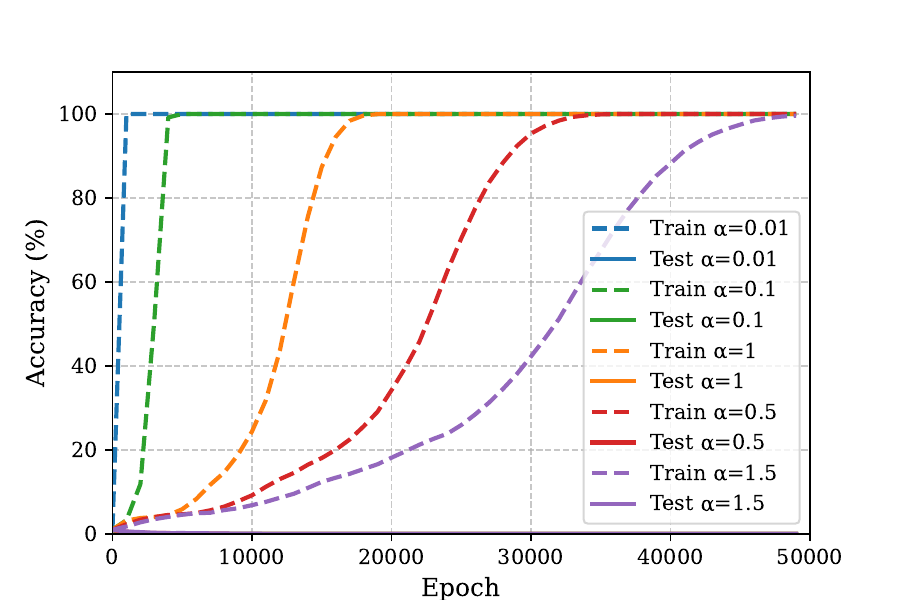}
  \caption{CE: 2 hidden layers}
\end{subfigure}
\caption{The $\alpha$ parameter controls generalization in settings where it happens by default. This is the case for shallow networks with MSE loss as shown in subplot (a). However, in deeper networks (b) or networks with CE loss and no regularization (c), $\alpha$ can control the time of over-fitting, but no value of $\alpha$ is enough to trigger grokking.}
\label{fig:alpha_parameter}
\end{figure}

\paragraph{Scaling the logits can delay generalization but not induce it} \cite{liu2023omnigrok}, \cite{Kumar2023-hz} and \cite{Lyu2023-ga} showed that an $\alpha$ parameter multiplying the logits can increase or reduce the delay in generalization. We highlight in \cref{fig:alpha_parameter} that this is true for cases where generalization happens even without changing the scale of the logits ($\alpha=1$). However, in most cases when using deeper networks or cross-entropy loss, models do not generalize by default without regularization and we are unable to induce grokking for any value of $\alpha$. 

We argue in \cref{sec:explain_existing_methods} that the observation in \cite{liu2023omnigrok}, \cite{Kumar2023-hz} and \cite{Lyu2023-ga} of grokking without regularization are due to the inductive bias of MSE loss which prevents NLM and leads to grokking in some settings for shallow networks.

\paragraph{Grokking on MNIST} We replicate the setting from \cite{liu2023grokking} of grokking on MNIST with cross-entropy loss and show that without weight decay, the scaling factor of the weights leads to significant FP errors, preventing grokking from happening until this is alleviated by weight decay. 

\begin{figure}[t]
    \vspace{-6mm}
    \centering
    \begin{subfigure}{.48\textwidth}
    \includegraphics[width=\linewidth]{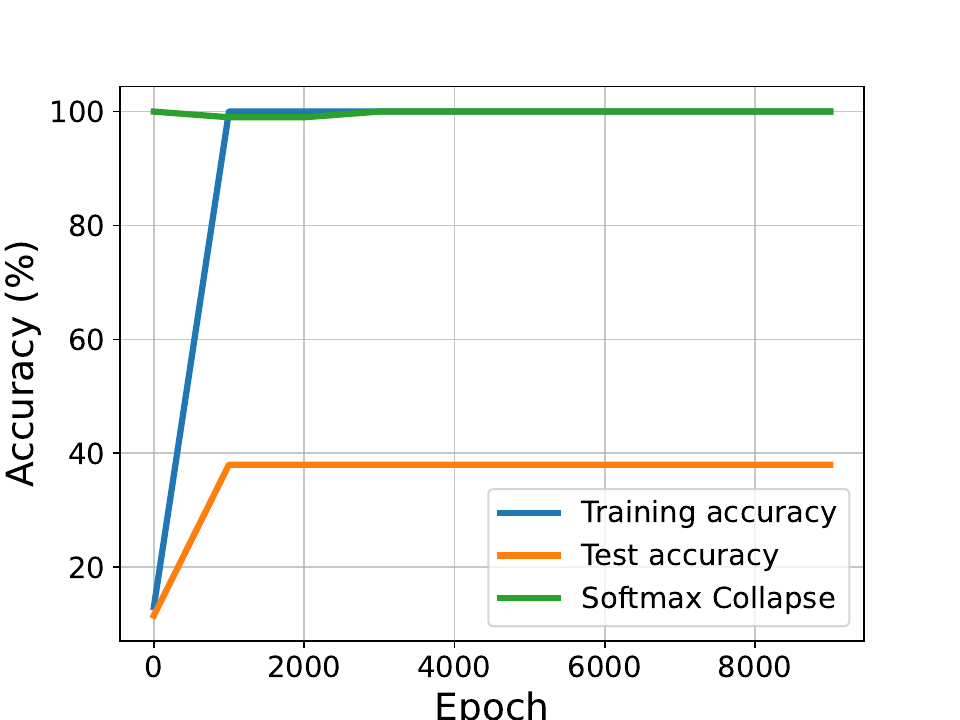}
        \caption{MLP without weight decay}
        \label{fig:mnist_witout_weight_decay}
    \end{subfigure}
    \begin{subfigure}{.48\textwidth}
    \includegraphics[width=\linewidth]{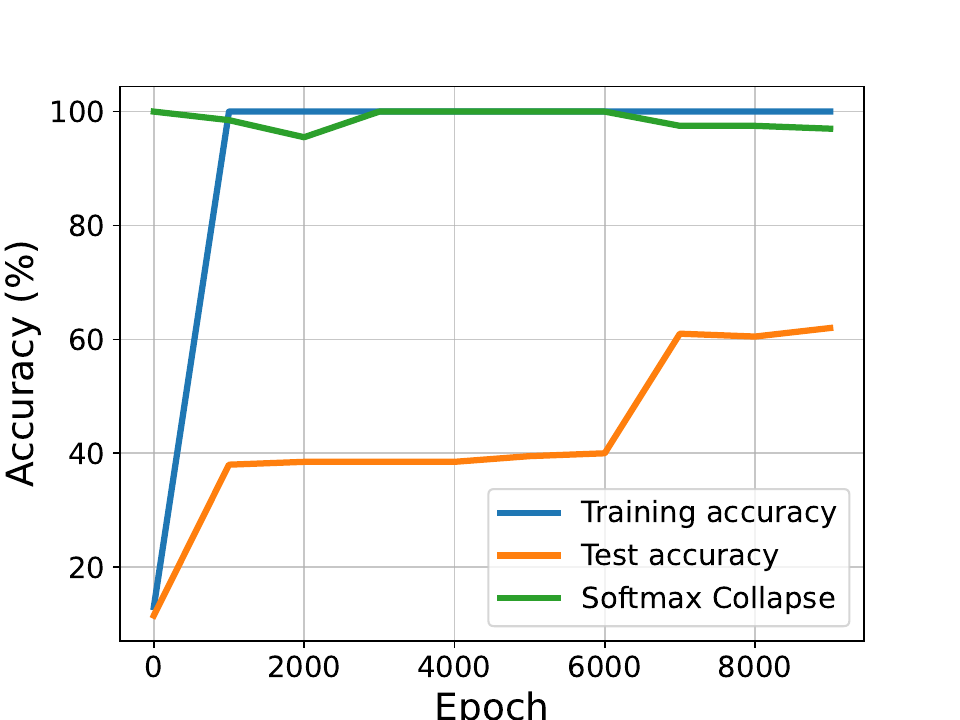}
        \caption{MLP with weight decay.}
        \label{fig:mnist}
    \end{subfigure}
    \caption{Replicating the grokking on MNIST for weight decay setting from \cite{liu2023grokking}. We find that MLPs with weights scaled up by 100 operate at the ``edge of numerical stability'' and in the absence of weight decay, SC eventually reaches 100\%, preventing any further generalization. When using weight decay, the weight norm is reduced, mitigating SC and eventually allowing for further generalization as the SC rate drops from 100\%.}

\end{figure}

While SC explains why weight decay is needed to get the jump in performance observed in \cref{fig:mnist}. It could also explain why inducing grokking by scaling the weights is less effective when using SCE. While when using MSE loss, \cite{liu2023omnigrok} are able to induce full grokking from random level predictions to close to full training accuracy, the same does not seem to be possible when using SCE. In fact, we see in \cref{fig:mnist} that since the beginning of training the rate of SC approaches 100\%. This could explain why the observations with cross-entropy loss are not the ones predicted by the lazy training theories outlined in \cite{Kumar2023-hz} which do not take limited floating point precision into account.

\section{\ograd and Weight Decay}
\begin{figure}[t]
    \vspace{-5mm}
    \centering
    \begin{subfigure}{.48\textwidth}
    \includegraphics[width=\linewidth]{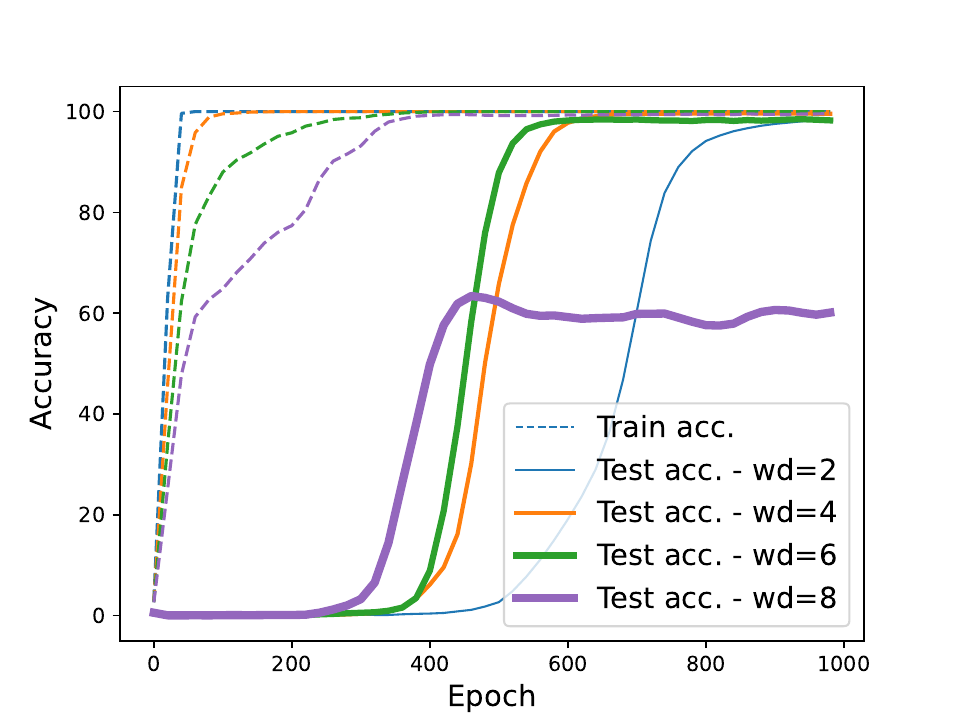}
        \caption{Sweep over values of weight decay}
        \label{fig:wd_sweep}
    \end{subfigure}
    \begin{subfigure}{.48\textwidth}
    \includegraphics[width=\linewidth]{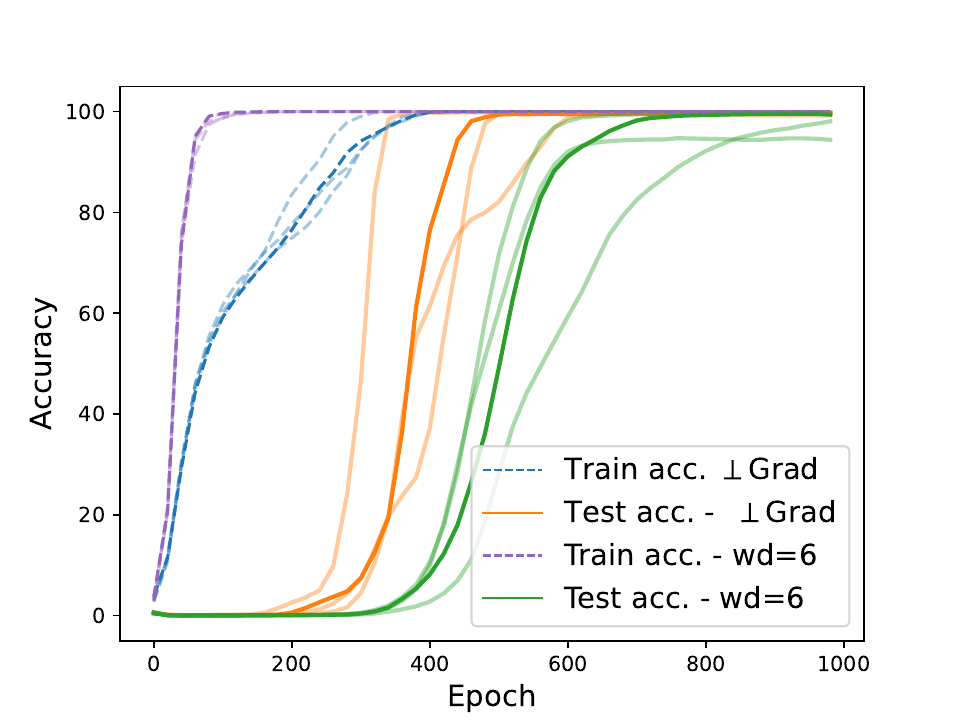}
        \caption{\ograd vs best performing wd model}
        \label{fig:vs_wd_seeds}
    \end{subfigure}
    \caption{Increasing weight decay (WD) for an MLP trained on modular addition with AdamW reduces the delay in generalization up to a point where WD prevents convergence \cref{fig:wd_sweep}. Without any tunable hyper-parameters and without WD, \ograd leads to grokking faster than the best model with WD \cref{fig:vs_wd_seeds}.}
    \label{fig:wd_vs_ortho}
\end{figure}

In \cref{fig:wd_vs_ortho}, we provide a more in depth comparison of \ograd and weight decay. \cref{fig:wd_sweep} highlights that increasing the weight decay multiplier leads to a smaller delay in generalization, but only up to a point. In this concrete setting, a weight decay multiplier of 8, prevents the model from fully generalizing (\cref{fig:wd_sweep}). We then compare the best value of weight decay in this setting to \ograd, which does not require any hyper-parameter tuning. \cref{fig:vs_wd_seeds} shows that \ograd leads to faster grokking even when compared to a tuned value of weight decay. Note that the models with weight decay overfit immediately before grokking while \ograd reaches 100\% train and test accuracies almost at the same time.

\section{Alternatives to $\stablemax$ in Preventing SC}
\begin{wrapfigure}[15]{R}{0.45\textwidth}
    \vspace{-7mm}
    \includegraphics[width=\linewidth]{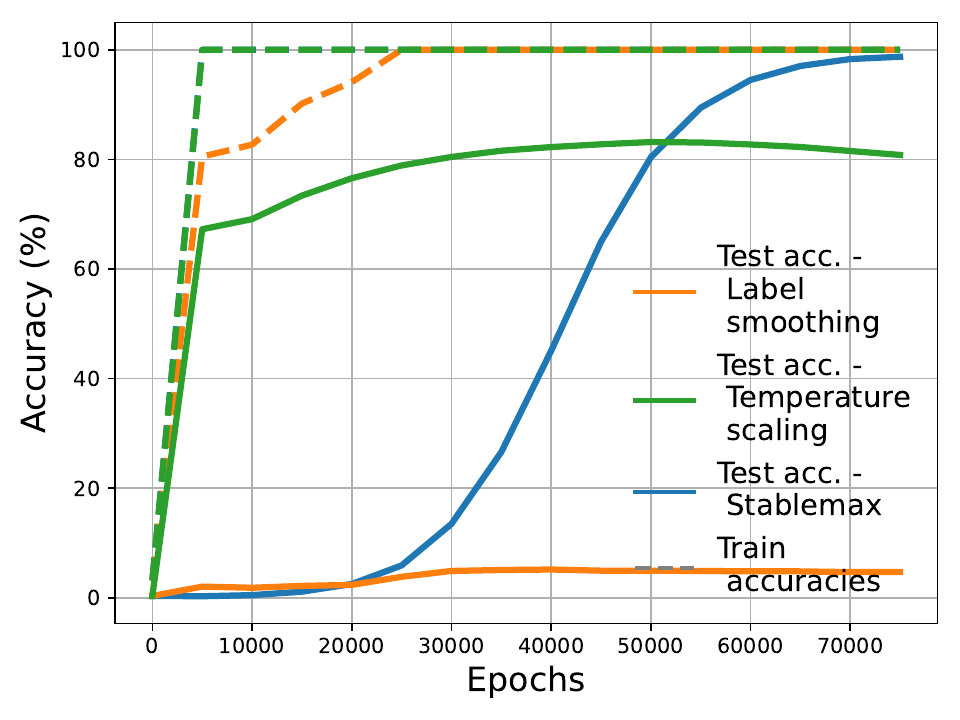}
    \vspace{-8mm}
    \caption{$\stablemax$ prevents SC and leads to grokking while temperature scaling with $T=1e5$ only gradually delays SC, and label smoothing does prevent SC but at the cost of keeping the model from fully generalizing.}
    \label{fig:label_smoothing}
\end{wrapfigure}
While any intervention that prevents SC should lead to grokking or generalization, \cref{fig:label_smoothing} shows that scaling the temperature of the \softmax is not enough to prevent SC and label smoothing does prevent SC and lead to some generalization, but at the cost of introducing another inductive bias that prevents full generalization and leads to qualitatively different behavior. By comparison, the simple change introduced in $\stablemax$ prevents SC and leads to grokking, serving as a validation for our hypothesis that gradient descent leads to grokking by default, unless this is stopped by SC.

\section{$\stablemax$ and \ograd \\ in Realistic Settings}
While $\stablemax$ and \ograd are designed as interventions to show that preventing SC leads to grokking and preventing NLM leads to generalization (\cref{fig:teaser}), in this section we explore if these methods are applicable in more realistic settings like language modeling with GPT2-small or ResNets trained on image classification. We train GPT2-Small for 1 epoch on WikiText-103 using a batch size of 16, a block size of 512, a learning rate of $5e-4$ and a weight decay of 0.01 using AdamW. The architecture is the regular GPT2-Small architecture from \cite{radford2019language}, trained with a cosine schedule and 1000 steps of warm-up.

For CIFAR10, CIFAR100 and Imagenet-1k \citep{ILSVRC15}, our baseline is a ResNet18 with SCE loss trained with SGD 0.9 momentum and $1e-4$ weight decay. We use standard data transformations such as random crop and random horizontal flip and a step learning rate scheduler every 30 epochs for a full training run of 100 epochs. With respect to this baseline we report results replacing the $\softmax$ with $\stablemax$ in the loss function, as well as replacing SGD with $\perp$SGD. Since test labels for Imagenet-1k are not publicly available, we use the validation set as a test set and tune hyper-parameters on a fraction of the training set.

\begin{figure}[t]
    \centering
    \begin{subfigure}{.32\textwidth}
    \includegraphics[width=\linewidth]{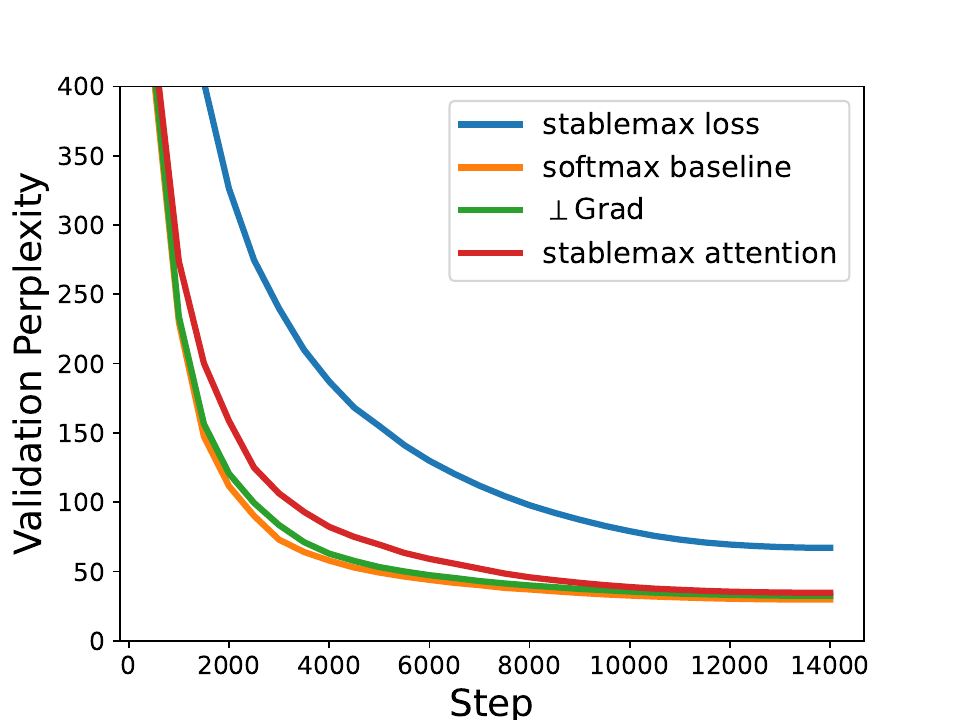}
        \caption{GPT2-Small on WikiText2}
        \label{fig:gpt2_small}
    \end{subfigure}
    \hfill
    \begin{subfigure}{.32\textwidth}
    \includegraphics[width=\linewidth]{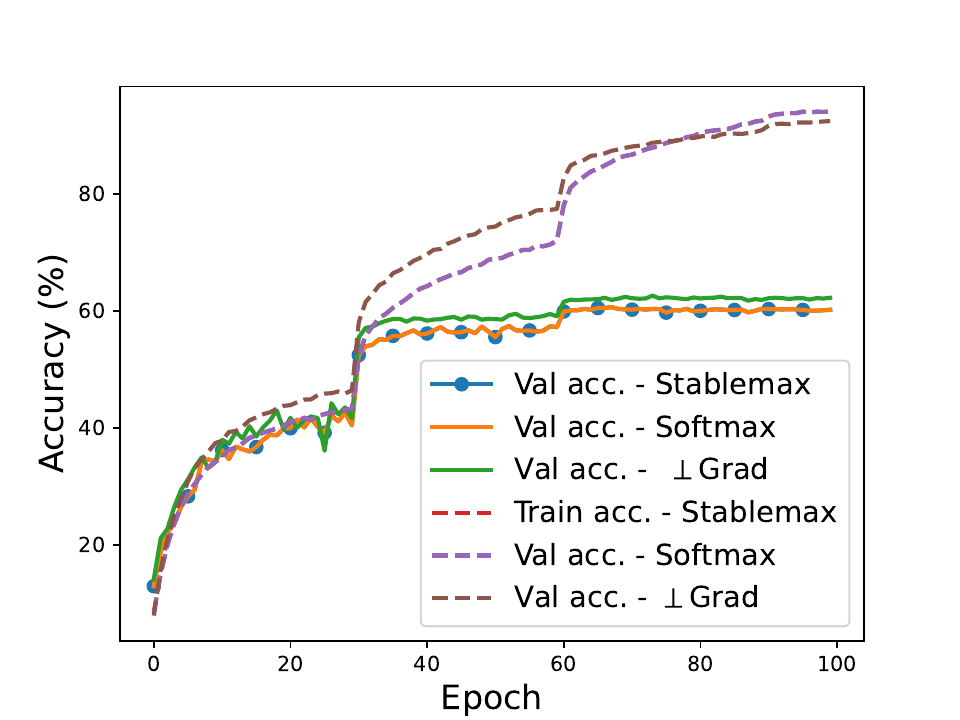}
        \caption{ResNet18 on CIFAR100}
        \label{fig:cifar100_resnet18}
    \end{subfigure}
    \hfill
    \begin{subfigure}{.32\textwidth}
    \includegraphics[width=\linewidth]{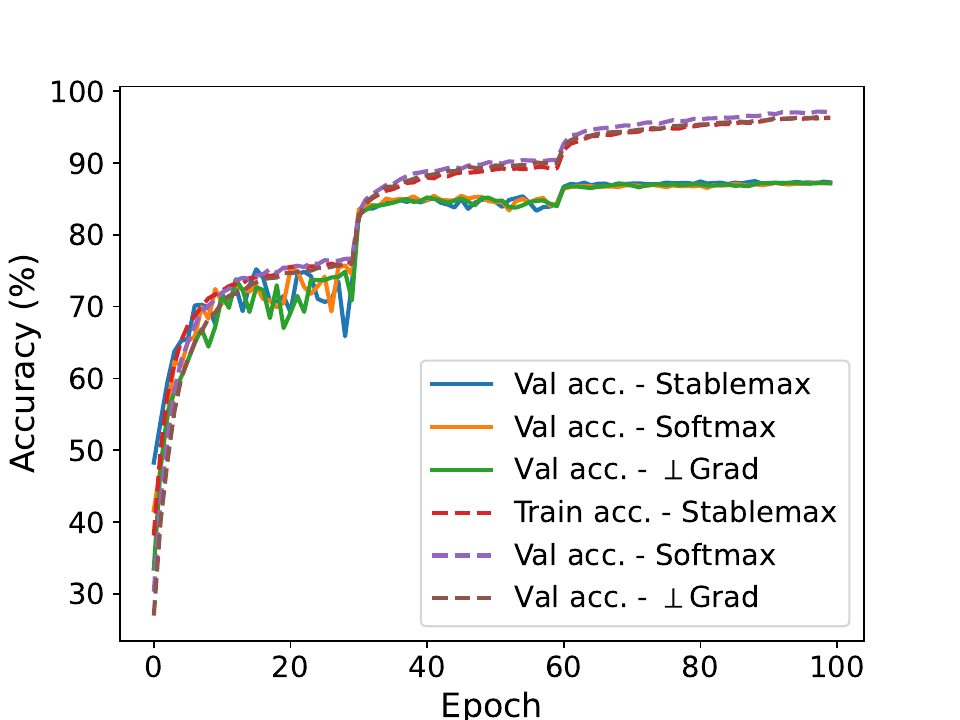}
        \caption{ResNet18 on CIFAR10}
        \label{fig:cifar10_resnet18}
    \end{subfigure}
    \caption{Comparing Stablemax and \ograd to AdamW with SCE on text data \cref{fig:gpt2_small} and image data \cref{fig:cifar10_resnet18}. For the GPT2-small results in \cref{fig:gpt2_small}, we also include the results of replacing the \softmax in the attention mechanism with $\stablemax$.}
\end{figure}

\begin{table}[h]
\centering
\begin{tabular}{@{}lcccc@{}}
\toprule
\textbf{Method}         & \textbf{CIFAR10} & \textbf{CIFAR100} & \textbf{ImageNet-1k} & \textbf{WikiText-103} (Top-5) \\ \midrule
Softmax CE              & $87.17\% \pm 0.2$ & $59.98\% \pm 0.4$  & $69.33\% \pm 0.04$              & $60.48\% \pm 0.04$                      \\
Stablemax CE            & $87.01\% \pm 0.2$ & $60.63\% \pm 0.4$  & $65.87\% \pm 0.22$             & $51.85\%  \pm 0.47$                  \\
\ograd                  & $87.22\% \pm 0.2$ & $62.69\% \pm 0.1$  & $68.95\% \pm 0.03$               & $59.64\%   \pm 0.04$                \\ 
\midrule
Stablemax Attention     & --                & --                 & --                   & $58.52\%   \pm 0.04$                   \\ \bottomrule
\end{tabular}
\label{tab:realistic_datasets}
\caption{For the methods introduced in this paper, we report accuracies with standard deviations across five seeds for the CIFAR datasets and three seeds for Imagenet-1k and WikiText-103. We report Top-5 accuracy in the case of WikiText-103.\vspace{-3mm}}
\end{table}

\section{SC and the Slingshot Effect}\label{app:slingshots}
\cite{slingshot-mechanism} observed that spikes in the training loss appear when training on grokking tasks with adaptive optimizers like Adam, and that these spikes can lead to generalization without weight decay. Although \cite{Nanda2023-hf} showed that slingshots are not necessary for grokking, it is still unclear what mechanism of adaptive gradient optimizers induces this behavior and why it leads to generalization. In light of the results in this paper, we believe that slingshots could lead to generalization because they prevent full SC. \cite{Nanda2023-hf} pointed out that something like SC could be responsible for these slingshots. One possible mechanism would be that zero gradients for some samples due to SC rapidly diminish the second-order moments leading to a large update or slingshot which moves the model away from full SC, although more research would be needed to properly show this.

While related to our work, slingshots are a different kind of instability which only appears with adaptive optimizers and can allow grokking. In contrast, we identify SC as a very specific issue in the \softmax that can affect any model trained with SCE, not only the ones trained with adaptive optimizers. Additionally SC prevents grokking whereas slingshots can lead to it. Wether and how slingshots are cause by SC remains an open research question, with some supporting evidence from \cite{Nanda2023-hf} which show that slingshots can disappear when using $float64$.

\section{Additional Details About Floating Points}
Beyond our main results, we found that in some cases, grokking could be stopped before SC due to the $\epsilon$ parameter in Adam being too large. While the $\epsilon$ term is designed to give numerical stability to the gradients, in settings with extremely low losses and gradients, the second order moments can be dominated by the $\epsilon$ term, putting an end to learning where it would have continued with a smaller $\epsilon$ value. This echoes the results in \cite{slingshot-mechanism} which shows that increasing $\epsilon$ halts slingshots and grokking, with \cite{Nanda2023-hf} also alluding to the $\epsilon$ parameter being important in some cases.  

Surprisingly, we also found that a simple re-implementation of $torch.nn.functional.log\_softmax$ that does not use the official CUDA kernels can lead the models to keep learning beyond the point where the loss is exactly 0 and some gradients should be 0 with appropriate calculation, outperforming the official implementation for grokking tasks. Learning eventually also stops in this setting and this seems more like a quirk of how gradients are calculated in PyTorch in the absence of an explicitly defined backward pass.

\end{document}